\pdfoutput=1
\documentclass[11pt]{article}

\usepackage[]{acl}
\usepackage{times}
\usepackage{latexsym}
\usepackage{booktabs}
\usepackage{booktabs}
\usepackage{amsfonts}
\usepackage{nicefrac}
\usepackage{microtype}
\usepackage{todonotes}
\usepackage{listings}
\usepackage{algorithm}
\usepackage{algpseudocode}
\usepackage{soul}
\usepackage{tabularx}
\usepackage{pifont}
\usepackage{xcolor}
\usepackage{graphicx}
\usepackage{subcaption}
\usepackage{hyperref}
\usepackage{comment}
\usepackage{xspace}
\usepackage{dsfont}
\usepackage{colortbl}
\usepackage{xcolor}
\usepackage{makecell}
\usepackage{multirow}

\usepackage[normalem]{ulem}
\definecolor{green}{RGB}{36, 214, 36}
\definecolor{red}{RGB}{235, 30, 30}
\definecolor{lightredshade}{HTML}{dea9a9}
\definecolor{lightgreenshade}{HTML}{bce3bd}
\definecolor{lightblueshade}{HTML}{cacbe8}
\definecolor{MyDarkBlue}{rgb}{0,0.08,1}
\definecolor{MyDarkGreen}{rgb}{0.02,0.6,0.02}
\definecolor{MyDarkRed}{rgb}{0.8,0.02,0.02}
\definecolor{MyDarkOrange}{rgb}{0.40,0.2,0.02}
\definecolor{MyPurple}{RGB}{111,0,255}
\definecolor{MyRed}{rgb}{1.0,0.0,0.0}
\definecolor{MyGold}{rgb}{0.75,0.6,0.12}
\definecolor{MyDarkgray}{rgb}{0.66, 0.66, 0.66}

\definecolor{MyYellow}{rgb}{254, 246, 170}
\definecolor{MyBlue}{rgb}{170, 217, 251}
\definecolor{LuneBlue}{rgb}{0.11, 0.11, 0.43}

\newcommand{\greencheck}{\textcolor{green}{\ding{51}}}
\newcommand{\redcross}{\textcolor{red}{\ding{55}}}

\definecolor{darkgreen}{RGB}{0,100,0}

\newcommand{\eg}{{\it e.g.},~}%
\newcommand{\ie}{{\it i.e.},~}%

\newcommand{\coffeegym}{\textsc{Coffee-Gym}\xspace}
\newcommand{\cf}{\textsc{Coffee}\xspace}
\newcommand{\coffeeeval}{\textsc{CoffeeEval}\xspace}
\newcommand{\editeval}{\textsc{CoffeeEval}\xspace}
\newcommand{\editevalbf}{\textbf{\textsc{CoffeeEval}}\xspace}
\newcommand{\coffeewemoji}{\coffee\xspace\cf}
\newcommand{\coffeewemojibf}{\coffee\xspace\textbf{\cf}}
\newcommand{\cfwemoji}{\coffee\xspace\cf}

\definecolor{pythonblue}{rgb}{0.16,0.12,0.93}
\definecolor{cppgreen}{rgb}{0.16,0.42,0.16}
\definecolor{promptinsert}{HTML}{bfefff}
\definecolor{compcolor}{HTML}{90EE90}
\definecolor{codehlcolor}{HTML}{ffec8b}
\definecolor{codehlcolor2}{HTML}{ffbbff}
\definecolor{bgcolor}{rgb}{0.95,0.95,0.92}

\lstdefinestyle{python}{
    language=Python,
    basicstyle=\fontsize{8}{10}\ttfamily,
    keywordstyle=\color{blue},
    commentstyle=\color{gray},
    stringstyle=\color{black},
    showstringspaces=false,
    breaklines=true,
    breakindent=0pt,
    breakatwhitespace=false,
    escapeinside={(*@}{@*)}
}

\lstdefinestyle{cpp}{
    language=C++,
    basicstyle=\fontsize{8}{10}\ttfamily,
    keywordstyle=\color{blue},
    commentstyle=\color{gray},
    stringstyle=\color{green},
    showstringspaces=false,
    breaklines=true,
    breakindent=0pt,
    breakatwhitespace=false,
    escapeinside={(*@}{@*)}
}

\lstdefinestyle{plain}{
    basicstyle=\fontsize{8}{10}\ttfamily,
    keywordstyle=\color{blue},
    commentstyle=\color{gray},
    stringstyle=\color{green},
    showstringspaces=false,
    breaklines=true,
    breakatwhitespace=false,
    breakindent=0pt,
    escapeinside={(*@}{@*)}
}

\lstdefinestyle{python2}{
    language=Python,
    basicstyle=\fontsize{8}{10}\ttfamily,
    keywordstyle=\color{blue},
    commentstyle=\color{gray},
    stringstyle=\color{green},
    showstringspaces=false,
    breakatwhitespace=false,
    breaklines=true,
    breakindent=0pt,
    escapeinside={(*@}{@*)}
}

\lstdefinestyle{cpp2}{
    language=C++,
    basicstyle=\fontsize{8}{10}\ttfamily,
    keywordstyle=\color{blue},
    commentstyle=\color{gray},
    stringstyle=\color{green},
    showstringspaces=false,
    breaklines=true,
    breakindent=0pt,
    breakatwhitespace=false,
    escapeinside={(*@}{@*)}
}

\lstdefinestyle{sql}{
    language=SQL,
    basicstyle=\fontsize{8}{10}\ttfamily,
    keywordstyle=\color{blue},
    commentstyle=\color{green},
    stringstyle=\color{black},
    showstringspaces=false,
    breakatwhitespace=false,
    breaklines=true,
    breakindent=0pt,
    escapeinside={(*@}{@*)}
}

\lstdefinestyle{prompt}{
    language=Python,
    basicstyle=\fontsize{8}{10}\ttfamily,
    keywordstyle=\color{blue},
    commentstyle=\color{gray},
    stringstyle=\color{cppgreen},
    showstringspaces=false,
    breaklines=true,
    backgroundcolor=\color{bgcolor},
    keepspaces=true, 
    breakindent=0pt,
    breakatwhitespace=false,
    showspaces=false,   
    escapeinside={(*@}{@*)}
}
\lstdefinestyle{text}{
    basicstyle=\fontsize{8}{10}\ttfamily,
    showstringspaces=false,
    breaklines=true,
    backgroundcolor=\color{bgcolor},
    breakatwhitespace=false,
    breakindent=0pt,
    keepspaces=true,
    showspaces=false,   
    escapeinside={(*@}{@*)}
}

\definecolor{lightblue}{RGB}{224,236,247}
\definecolor{deepblue}{RGB}{9,46,107}

\usepackage[T1]{fontenc}
\usepackage[utf8]{inputenc}

\usepackage{microtype}

\usepackage{hyperref}
\usepackage{multirow}
\usepackage{graphicx}
\usepackage{pdfpages}
\usepackage{url}
\usepackage{xcolor}
\usepackage{epsfig}
\usepackage{adjustbox}
\usepackage{amsfonts}
\usepackage{amsmath}
\usepackage{amssymb}
\usepackage{booktabs} 
\usepackage{comment}
\usepackage{caption}
\usepackage{subcaption}
\usepackage{textcomp}
\usepackage{relsize}
\usepackage{stmaryrd}
\usepackage{bbm}
\usepackage{rotating}
\usepackage{helvet}
\usepackage{courier}
\usepackage{natbib}
\usepackage{cleveref}
\usepackage{arydshln}
\usepackage{listings}
\usepackage{wrapfig}
\usepackage{longtable}
\usepackage{adjustbox}
\usepackage{listings}
\usepackage{tcolorbox}
\tcbuselibrary{raster}
\tcbuselibrary{breakable}
\usepackage{lscape}
\definecolor{Box2Color}{rgb}{0.95,0.95,0.95}
\PassOptionsToPackage{table}{xcolor}

\newcommand\coffee{\raisebox{-2pt}{\includegraphics[width=1em]{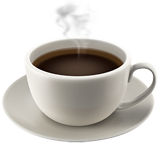}}}

\crefformat{section}{\S#2#1#3} % see manual of cleveref, section 8.2.1

\title{\textsc{Coffee-Gym}: An Environment for Evaluating and Improving \\Natural Language Feedback on Erroneous Code}

\author{\qquad Hyungjoo Chae$^{1*}$\qquad Taeyoon Kwon$^{1*}$\qquad Seungjun Moon$^{1}$\thanks{~~Equal contribution}\qquad \\ 
\textbf{Yongho Song}\qquad\textbf{Dongjin Kang}$^{1}$\qquad\textbf{Kai Tzu-iunn Ong}$^{1}$\qquad\textbf{Beong-woo Kwak}$^{1}$\\ \textbf{Seonghyeon Bae}$^{1}$ \qquad \textbf{Seung-won Hwang}$^{2}$ \qquad \textbf{Jinyoung Yeo}$^{1}$ \\ 
Yonsei University$^{1}$  \quad Seoul National University$^{2}$\\
\texttt{\{mapoout, kwonconnor101, lune\_blue, jinyeo\}@yonsei.ac.kr} \\ \texttt{seungwonh@snu.ac.kr}\\
}

\begin{document}
\maketitle
\begin{abstract}
This paper presents \coffeegym, a comprehensive RL environment for training models that provide feedback on code editing. \coffeegym includes two major components: (1) \cf, a dataset containing humans' code edit traces for coding questions and machine-written feedback for editing erroneous code; (2) \coffeeeval, a reward function that faithfully reflects the helpfulness of feedback by assessing the performance of the revised code in unit tests.
With them, \coffeegym addresses the unavailability of high-quality datasets for training feedback models with RL, and provides more accurate rewards than the SOTA reward model (\ie GPT-4). By applying \coffeegym, we elicit feedback models that outperform baselines in enhancing open-source code LLMs' code editing, making them comparable with closed-source LLMs. We make the dataset and the model checkpoint publicly available.\footnote{\url{https://huggingface.co/spaces/Coffee-Gym/Project-Coffee-Gym}}  
\end{abstract}

\section{Introduction}

Large language models (LLMs) have made great progress in code generation~\citep{li2023starcoder,roziere2023code}, \eg achieving human-level performances in code generation benchmarks~\citep{chen2021evaluating}. Such success makes them powerful tools for assisting human programmers~\citep{köpf2023oasst};
however, they still produce errors~\citep{Guo2024DeepSeekCoderWT, openai2023gpt4}.
Therefore, code editing, \ie resolving errors in code, remains an important task for code LLMs~\citep{ muennighoff2023octopack}.

\begin{figure}[t!]
\centering
    \includegraphics[width=0.9\linewidth]{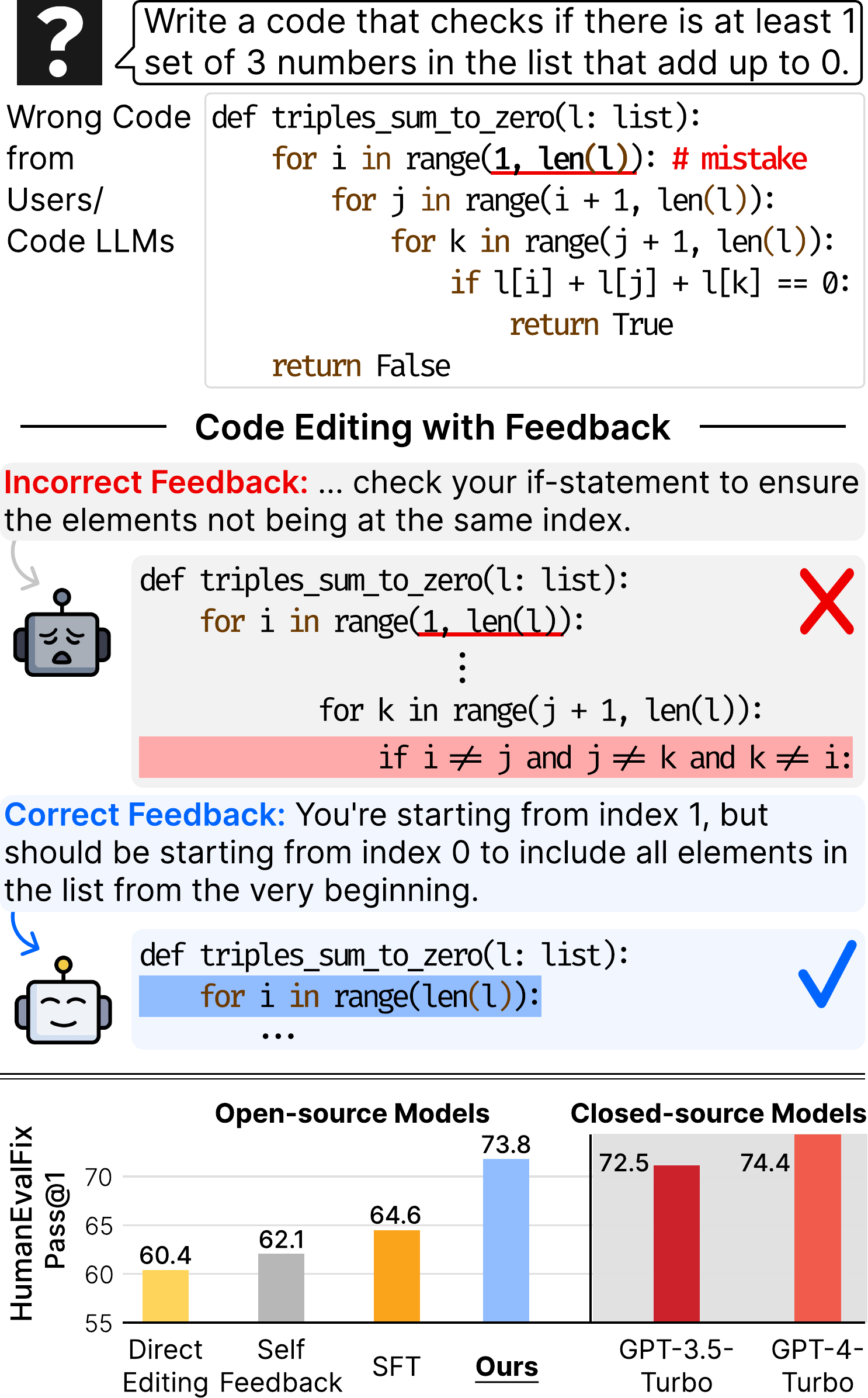}
\caption{A motivating example (Top) and Pass@1 accuracy in HumanEvalFix (Bottom). We compare the feedback from our model and various other models, both paired with DeepSeekCoder-7B as the code editor. SFT denotes the model trained on Code-Feedback~\citep{zheng2024opencodeinterpreter} using the same backbone model as ours.} 
\label{fig:motivate}
\end{figure}
\begin{figure*}[!h]
    \centering
    \includegraphics[width=1\linewidth]{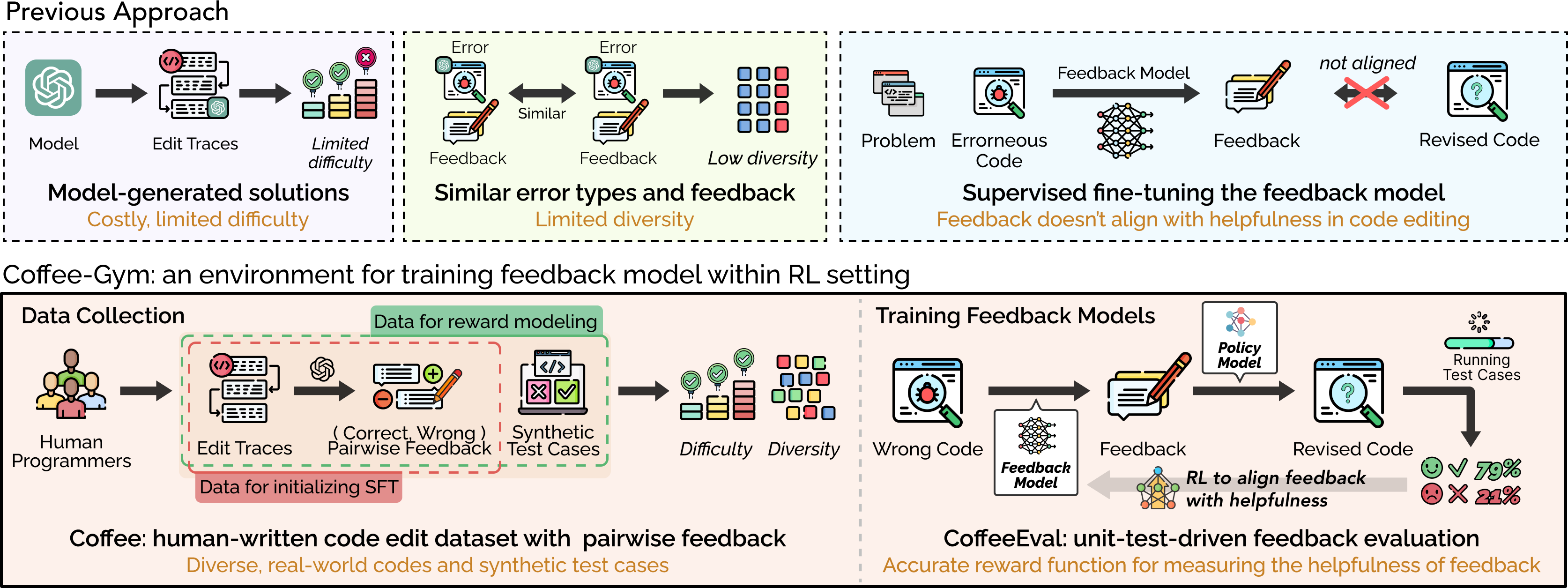}
    \caption{ Comparison between \coffeegym and the previous approach.}
    \label{fig:comparison_w_prev}
\end{figure*}

Studies have utilized natural language (NL) feedback from LLMs as descriptive guidance in editing wrong codes for code LLMs. For instance, Self-Refine~\citep{madaan2023selfrefine} largely improves their code editing using GPT-4's feedback. 
Yet, abilities to generate helpful feedback, as they report, are limited to powerful closed-source LLMs (\eg GPT-4).
This can lead to a heavy reliance on closed-source LLMs that may cause not only high computational (\eg API) cost but also security risks~\citep{Siddiq2023GenerateAP, Greshake2023NotWY}, limiting their applicability for confidential codes.

This work aims to foster building open-source feedback models that produce effective feedback for code editing.
An intuitive approach is to apply supervised fine-tuning (SFT) on open-source code LLMs using feedback from GPT-4 (generated based on machines' code editing)~\citep{zheng2024opencodeinterpreter}. 
However, this simplified approach poorly aligns editing performance with the helpfulness of feedback (Bottom of Figure~\ref{fig:motivate})~\citep{liu2022rainier}.

Inspired by the success of RLHF~\citep{Ouyang2022TrainingLM}, we reformulate feedback modeling with reinforcement learning (RL), where we align feedback models with the helpfulness of feedback during training.
% Inspired by the recent success of RLHF~\citep{Ouyang2022TrainingLM}, we reformulate feedback modeling task into reinforcement learning (RL) setting, where our goal is to align feedback models with helpfulness on code editing during training.
Since the success of RL highly depends on the initial SFT model and a reliable reward function~\citep{lightman2023let,lambert2024rewardbench},
% Prior studies~\citep{lightman2023let,lambert2024rewardbench} have shown that the success of training models with RL mostly depends on the capability of initial SFT model and reliability of reward function.
we hereby identify 3 main challenges in applying RL to feedback generation for code editing: \textbf{(1)} limited scenarios of errors in model-generated code editing datasets for initializing SFT model, \textbf{(2)} the lack of pairwise (correct and wrong) feedback to train/test reward functions, \textbf{(3)} absence of validated implementation of reward models.

We present \textbf{\coffeegym}, a comprehensive RL environment addressing the above challenges in training feedback models for code editing.
First, to tackle data scarcity in SFT initialization and reward modeling, we curate \textbf{\coffeewemoji}, a dataset for \underline{co}de \underline{f}ixing with \underline{fee}dback, which consists of code editing traces of human programmers and human annotated feedback.
Unlike model-generated data (Figure~\ref{fig:comparison_w_prev}), \textsc{Coffee} includes (1) problems across various difficulties, including those current LLMs (\eg GPT-4) cannot solve; (2) pairs of correct and wrong feedback for reward modeling; (3) about 36 test cases per problem to measure the feedback helpfulness in code editing.\footnote{This work is a substantially revised and extended version of our preprint~\citep{moon2023coffee}. While both works use the same dataset, this submission presents significant advancements in methodology, analysis, and results.

}

Next, to address the absence of validated (\ie reliable) reward functions, we introduce \editeval, a reward function designed to reflect the helpfulness of feedback into reward calculation.
Instead of directly assessing feedback quality~\citep{rajakumar-kalarani-etal-2023-lets}, we simulate code editing based on generated feedback, conduct unit tests on the edited code, and use the test results to measure feedback helpfulness.
With the pairwise feedback from \coffeewemoji, we train a given code editor to produce edited code that faithfully reflects the helpfulness of the given feedback.

Through experiments, we validate \coffeegym's efficacy in training feedback models.
We find that \editeval provides more accurate rewards, compared to the current SOTA reward model, \ie G-Eval~\citep{liu2023geval} with GPT-4.
Also, we show that the feedback models trained with \coffeegym generate more helpful feedback, achieving comparable performance to closed-source feedback models in code editing.

\section{Task Definition and Problem Statement}

\subsection{Code Editing with Natural Language Feedback}
The task of code editing aims to resolve errors in given codes to produce a correct solution. Formally, given a problem description $q$ and a defective solution $y$, our goal is to learn a feedback model $\theta$ that generates helpful feedback describing the errors in $y$ and provide helpful guidance on code editing: $\hat{c} = \theta(q,y)$. Then, an editor model $\phi$ that takes $q$, $y$, and the generated feedback $\hat{c}$ as input and generates the edited code: $y' = \phi(q, y, \hat{c})$. 

\begin{figure*}[!t]
\centering
    \includegraphics[width=1\linewidth]{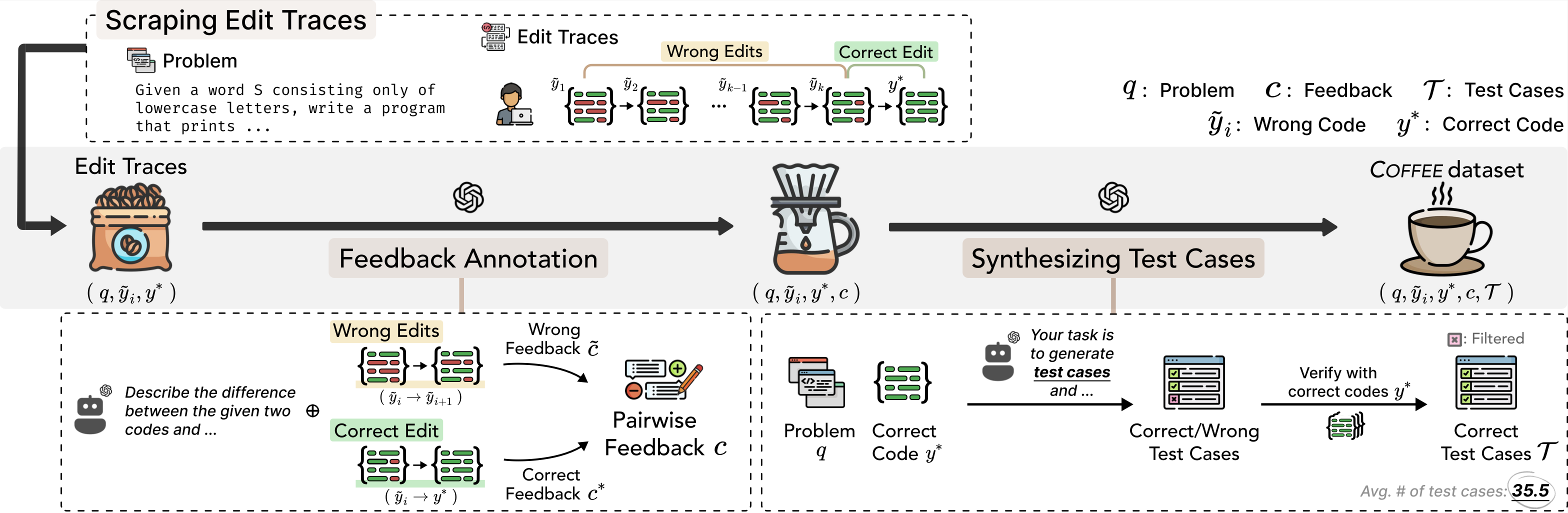}
\caption{Overview of the data collection process of \coffee~\cf.}
\label{fig:overview}
\end{figure*}

In evaluating the edited code $y'$, the functionality of the edited code is measured with Pass@k, the standard metric that measures the number of passed test cases $ t_i $ within the given set $ \mathcal{T} = \{t_1, t_2, \ldots, t_k\}$~\citep{li2022competition, li2023starcoder, muennighoff2023octopack}. Each test case $ t_i $ consists of an input $x_i$ and an expected output $z_i$.

\subsection{Learning Feedback Models}
In this paper, we consider two widely used learning approaches to build open-source feedback models. 
\paragraph{Supervised fine-tuning.} A straightforward approach is to fine-tune an open-source code LLM $\theta$ on a dataset $D = \{(q_i, y_i, c_i, y_i^*)\}_{i=1}^N$ of problem descriptions, incorrect codes, feedback annotations, and correct codes. 
The objective is to minimize the negative log-likelihood of the target feedback label $y^*$ given $q$ and $y$.
% \begin{equation}
% \mathcal{L}_{\text{SFT}}(\theta) = - \frac{1}{N} \sum_{i=1}^N \log p_\theta(c_i \mid q_i, y_i)
% \end{equation}
However, simply training to optimize the probability of the target sequence does not achieve much improvement for code editing, because it does not consider the impact of feedback on code editing~\citep{liu2022rainier}.

\paragraph{Reinforcement learning.} 
% DO NOT REMOVE =========================
% Recent advances of LLMs are largely based on reinforcement learning (RL)~\citep{bai2022training}. For example, RLHF is utilized to align LLMs with human preference~\citep{ziegler2019fine, Ouyang2022TrainingLM}.
% Inspired by this line of work, 
% DO NOT REMOVE =========================
Inspired by \citet{Ouyang2022TrainingLM}, we adopt reinforcement learning (RL) to further align feedback generation to correct code editing.
Specifically, we choose PPO~\citep{schulman2017proximal} and DPO~\citep{rafailov2023direct} as reference RL algorithms and apply them on the feedback model $\theta$ initialized via SFT.

The two key factors of RL are (1) \textbf{pairwise preference data} and (2) \textbf{reward modeling}~\citep{lambert2024rewardbench}. In our task, we consider a preference dataset where each input $q$ and $y$ comes with a pair of chosen and rejected feedback $c^+$ and $c^-$, and their preference ranking $c^+ \succ c^-$. This dataset is then used to model the reward based on the preference ranking. While in PPO a reward model is explicitly trained using $c^+$ and $c^-$, DPO relies on implicit reward modeling and directly optimizes the feedback model using the preference dataset.

\subsection{Problem Statement}
Our goal is to promote rapid development of open-source feedback models by facilitating RL for feedback generation on code editing. Specifically, we aim to provide the two key components in RL for feedback generation:

\paragraph{Dataset.} 
The dataset required for our RL approach covers the following key aspects: (1) \textbf{Coverage of difficulty and diversity ($q,y$)} to initialize a good SFT model.
(2) \textbf{Pairwise feedback data ($c^+ \succ c^- \mid q, y$)} to build datasets for training DPO and a reward model for PPO. 
(3) \textbf{Test cases for unit test ($\mathcal{T}$)} are required to implement our $R$, for directly measuring the impact of $c$ on the correctness of code editing.

\paragraph{Reward model.} 
% Training feedback models with RL algorithms requires a reward function that accurately measures the effectiveness of the generated feedback.
The current standard of using LLM as a reward model~\citep{lee2023rlaif} to evaluate LLM outputs do not sufficiently models the impact of feedback on code editing outcomes and requires powerful LLMs (\eg GPT-4) that incur high API costs.
Especially, the high computation costs significantly limits the application of online RL algorithms (\eg PPO) in feedback modeling, which require frequent and continuous API calls for reward calculation. 

\section{Constructing \coffeegym}
\label{sec:dataset}
We introduce \textsc{Coffee-Gym}, a comprehensive RL environment for training NL feedback model for code editing. 
\coffeegym consists of two major components: (1) \coffeewemojibf, a dataset of human-written edit traces with annotated NL feedback, and (2) \editevalbf, an accurate reward model that measures feedback's impact on code editing.

\begin{figure}[t]
\centering
    \includegraphics[width=0.9\linewidth]{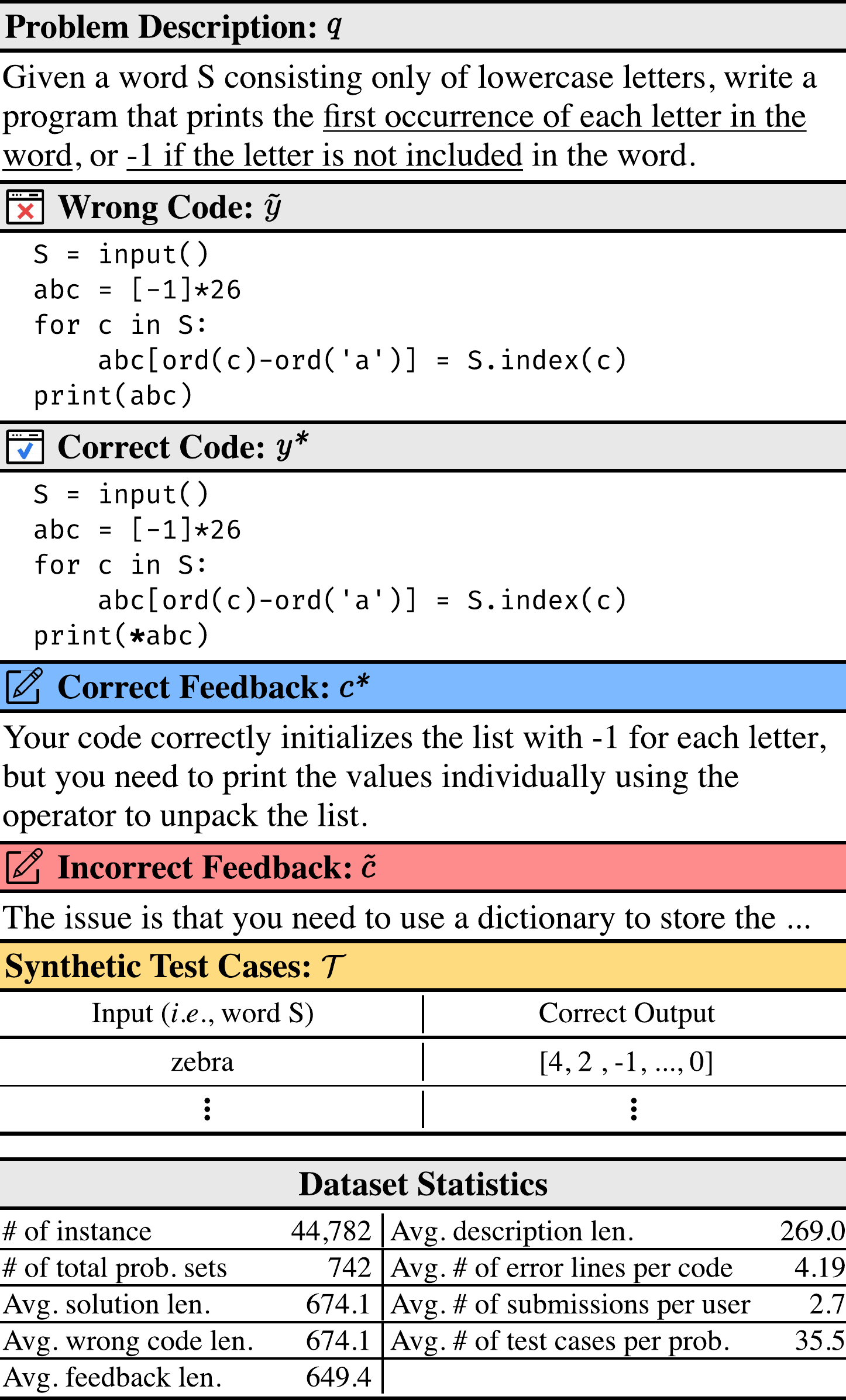}
\caption{Example and statistics of \coffeewemoji.} 
\label{fig:coffee_example}
\end{figure}

\subsection{\coffeewemoji: Human-written Code Edit Traces with Annotated Pairwise Feedback}
\label{ssec:coffee}
We curate \textbf{\cf}, a dataset of \underline{co}de \underline{f}ixing with \underline{fe}edback, from human-written code edit traces.
\cf consists of problems of diverse levels of difficulty, including challenging problems that only human programmers can solve, and provides test cases for reward functions (Section~\ref{ssec:coffeeeval}).
The overview of constructing \cf, data examples, and statistics are in Figure~\ref{fig:overview} and~\ref{fig:coffee_example}.

\subsubsection{Collecting Code Edit Traces from Human Programmers}
We collect human-authored code edits from an online competitive programming platform.\footnote{\url{https://www.acmicpc.net/}}
% To collect code edit data that include diverse error cases from human developers,
% we choose an online competitive programming platform as our data source.\footnote{\url{https://www.acmicpc.net/}}
In this platform, given a problem description $q$, human programmers keep submitting a new solution $y$ until they reach a correct solution $y^*$ that passes all hidden test cases for $q$.
Formally, for each $q$ and the correct submission $y_n^*$, we collect the submission history $\{\Tilde{y}_1, \Tilde{y}_2, ..., y^*_n\}$, where $\{\Tilde{y}_k\}_{k=1}^{n-1}$ are incorrect solutions. We then construct $(q,\Tilde{y},y^*)$ triplets by pairing each incorrect solution $\Tilde{y}_k$ with the correct one $y^*_n$, \ie {$\{(q,\Tilde{y}_{k}, y^*_n)\}_{k=1}^{n-1}$}.

\begin{figure}[t]
\centering
    \includegraphics[width=1\linewidth]{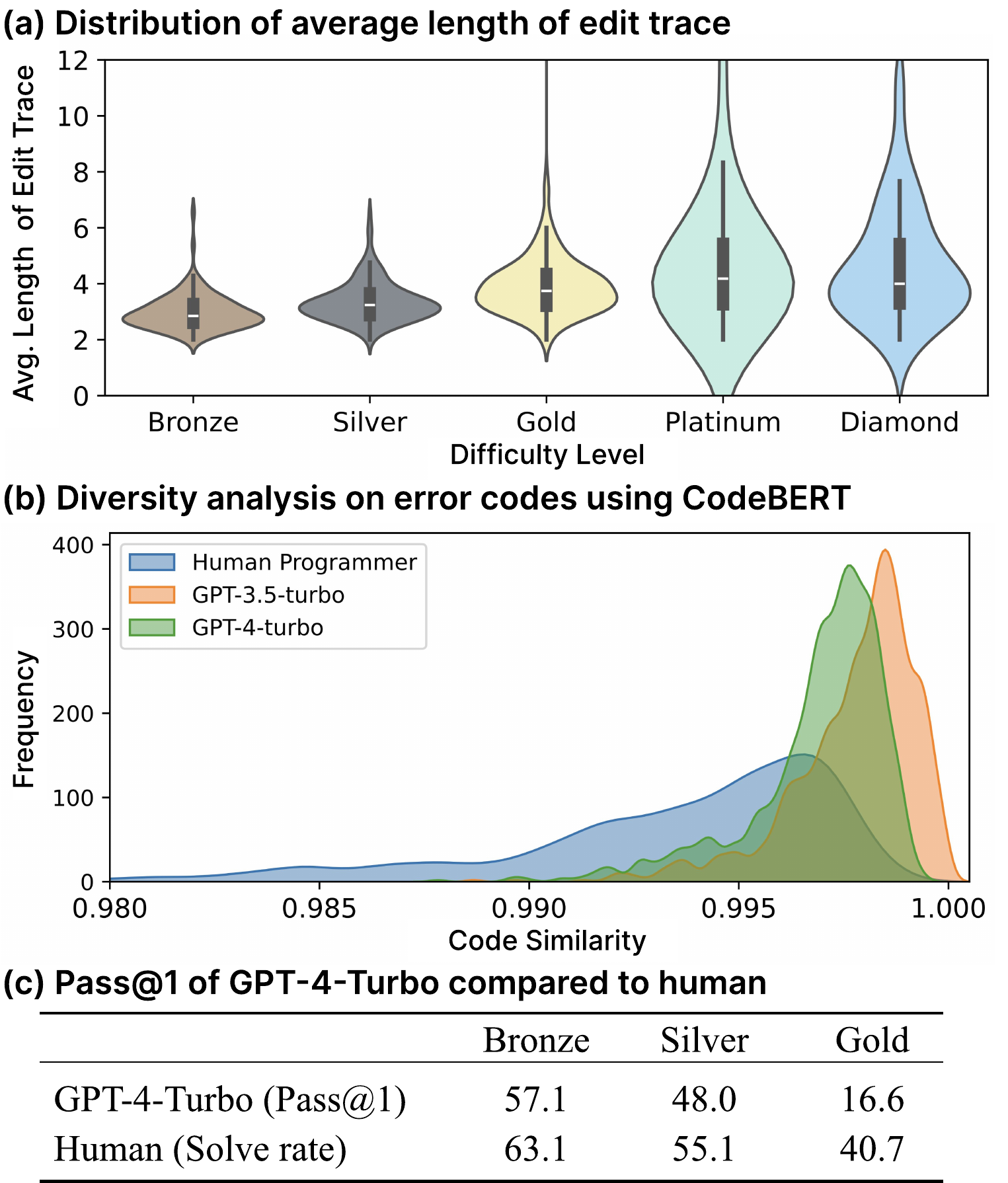}
\caption{Analysis results of \coffeewemoji. Experiment details are in Appendix~\ref{appendix:error_code_analysis}.}
\label{fig:violin}
\end{figure}

To ensure \cf is not biased toward coding problems of a specific difficulty level, we collect an equal number of problems from each of the five difficulty levels in the platforms, ranging from beginner to expert levels.
%Figure~\ref{fig:violin} shows that (1) humans tend to make more incorrect submissions on problems with higher difficulty levels and (2) the state-of-the-art LLM even struggles with the intermediate (\ie Gold) level.
%For generalized feedback modeling, 
We also ensure that \cf includes various solutions to each problem by collecting submission histories from 100 different users.
Our analysis in Figure~\ref{fig:violin} shows that \cf (1) includes problems that are challenging for both human and LLMs and (2) covers more diverse error cases than machine-generated codes.

\subsubsection{Annotating Pairwise Feedback Data}  
We additionally annotate NL feedback that provides useful guidance on the necessary edits. 
For each triplet $(q,\Tilde{y},y^*)$, we prompt GPT-3.5-Turbo~\citep{openai2023chatgpt} to describe how the correct solution $y^*$ differs from the wrong code $\Tilde{y}$.
%\footnote{We use \texttt{gpt-3.5-turbo-1106} model.} 
The resulting description $c^*$ serves as the correct feedback that describes necessary changes on the wrong code $\Tilde{y}$ to obtain the correct code $y^*$. 
Along with $c^*$, we also collect incorrect feedback $\Tilde{c}$, which describes the difference between two wrong solutions, $\Tilde{y}_{k-1}$ and $\Tilde{y}_{k}$ $(k \neq n)$, to provide pairwise labels for both correct and incorrect feedback to a single wrong solution $\Tilde{y}$.
% To ensure the quality of the annotated feedback, we exclude user submissions that do not involve error correction. 
We discuss details on feedback annotation in Appendix~\ref{appendix:feedback_annotation}, including our prompt used for feedback annotation and filtering techniques.

\subsubsection{Augmenting Synthetic Test Cases}
Finally, we include a set of hidden test cases $\mathcal{T} = \{ t_1, t_2, \dots, t_k \} $ for each edit instance $(q,\Tilde{y},y^*,c)$ in our dataset to assess whether the edited code is the correct solution to the problem. Each test case $t_i$ consists of an input $x_i$ and an expected output $z_i$. As the programming platform does not make test cases publicly available, we annotate test cases by prompting GPT-3.5-Turbo to generate inputs $x_i$ for a given $q$ and executing the correct code $y^*$ with $x_i$ to obtain the corresponding outputs $z_i$. We filter out any invalid test cases with inputs that result in errors during execution. On average, we obtain 35.5 test cases per problem.

\begin{table}[t]
% \small
% \renewcommand{\arraystretch}{1.15} %1.15
\centering

\resizebox{\columnwidth}{!}{
    \begin{tabular}{lccccccc}
    \toprule
    \textbf{}&\textbf{mean}&\textbf{std}&\textbf{min}&\textbf{25\%}&\textbf{50\%}&\textbf{75\%}&\textbf{max}\\
    \midrule
    Pass ratio & 0.342 & 0.370 & 0.000 & 0.000 & 0.162 & 0.693 & 0.985 \\
    
    \bottomrule
    \end{tabular}  
}

\caption{Pass ratio for incorrect code samples in the evaluation set of \textsc{Coffee} dataset.}
% The pass ratio is defined as the number of test cases passed divided by the total number of test cases.}
\label{tab:validity}
\end{table}

% \paragraph{Validity of test cases.}
A critical question in evaluating our test suite is whether any incorrect solutions manage to pass all the test cases. 
To address this, we conduct an experiment using the evaluation set of the \textsc{Coffee} dataset. 
We randomly sampled 200 wrong code instances and calculated the pass ratios of the wrong codes. 
We show the statistics of the distribution of pass ratios.
As shown in Table~\ref{tab:validity}, the maximum pass ratio is 0.985, which suggests that there are no wrong solutions that passed all the test cases. 
The mean score is 0.342, indicating that on average, wrong solutions fail the majority of the test cases. 
We further analyze the \textsc{Coffee-Test} and verified that no wrong solutions pass all the test cases.

These test cases are used to measure the correctness of an edited code and estimate the helpfulness of the feedback as the \editeval score, which we later use as supervision signals for training feedback models (\cref{ssec:coffeeeval}) in \coffeegym. We provide details on test case generation in Appendix~\ref{appendix:test_case}.

\begin{table*}[!t]
    \centering
    \resizebox{\textwidth}{!}{
    \begin{tabular}{lccc|ccc|c|c}
        \toprule
        \multirow{2}{*}{\textbf{Model}} & \multirow{2}{*}{\textbf{Evaluation}} & \multicolumn{2}{c}{\textbf{Pass@1}} & \multicolumn{3}{c}{\textbf{Scores}} & \textbf{Correlation}  & \textbf{Error}\\
        \cmidrule(lr){3-4} \cmidrule(lr){5-7} \cmidrule(lr){8-8} \cmidrule(lr){9-9}
        & & \greencheck~Correct Feedback $\uparrow$ (TP) & \redcross~Wrong Feedback $\downarrow$ (FP) & Precision $\uparrow$ & Recall $\uparrow$ & F1 $\uparrow$ & Pearson $\uparrow$ & MSE $\downarrow$ \\
        \midrule
        GPT-4-Turbo  & G-Eval & - & - & - & - & - & \cellcolor[HTML]{B6E4E7}\underline{0.135} & \cellcolor[HTML]{FC8A46}\underline{0.415} \\
        GPT-3.5-Turbo & G-Eval & - & - & - & - & - & \cellcolor[HTML]{F7FDFD}-0.172 & \cellcolor[HTML]{FFF6EF}0.575 \\
        \midrule
        GPT-4-Turbo & Editing & \cellcolor[HTML]{5BBFC6}\textbf{53.0} & \cellcolor[HTML]{FFF6EF}51.8 & \cellcolor[HTML]{D4F0F1}50.6 & \cellcolor[HTML]{5BBFC6}\textbf{53.0} & \cellcolor[HTML]{B6E4E7}\underline{51.8} & \cellcolor[HTML]{EBF8F9}0.012 & \cellcolor[HTML]{FED9BF}0.450 \\
        GPT-3.5-Turbo & Editing & \cellcolor[HTML]{B6E4E7}43.4 & \cellcolor[HTML]{FED9BF}33.6 & \cellcolor[HTML]{B6E4E7}\underline{56.4} & \cellcolor[HTML]{D4F0F1}43.4 & \cellcolor[HTML]{B6E4E7}49.0 & \cellcolor[HTML]{D4F0F1}0.101 & \cellcolor[HTML]{FC8A46}0.417 \\
        DeepSeek-Coder-7B & Editing & \cellcolor[HTML]{EBF8F9}36.0 & \cellcolor[HTML]{FDA66F}\underline{28.8} & \cellcolor[HTML]{D4F0F1}55.6 & \cellcolor[HTML]{EBF8F9}36.0 & \cellcolor[HTML]{D4F0F1}43.7 & \cellcolor[HTML]{D4F0F1}0.077 & \cellcolor[HTML]{FDA66F}0.428 \\
        DeepSeek-\editeval (w/o WF) & Editing & \cellcolor[HTML]{EBF8F9}36.4 & \cellcolor[HTML]{FB6522}\textbf{28.4} & \cellcolor[HTML]{B6E4E7}56.2 & \cellcolor[HTML]{EBF8F9}36.4 & \cellcolor[HTML]{D4F0F1}44.2 & \cellcolor[HTML]{D4F0F1}0.085 & \cellcolor[HTML]{FC8A46}0.418 \\
        DeepSeek-\editeval (Ours) & Editing & \cellcolor[HTML]{6bc5cc}\underline{52.0} & \cellcolor[HTML]{FB6522}\textbf{28.4} & \cellcolor[HTML]{5BBFC6}\textbf{64.7} & \cellcolor[HTML]{6bc5cc}\underline{52.0} & \cellcolor[HTML]{5BBFC6}\textbf{57.7} &  \cellcolor[HTML]{5BBFC6}\textbf{0.149} & \cellcolor[HTML]{FB6522}\textbf{0.408} \\
        \bottomrule
    \end{tabular} 
    }
    \caption{Performance of our evaluation protocol on the test sets of \cf compared to the baselines. Wrong Feedback is abbreviated as WF due to limited space.}
    \label{tab:evaluator_validation}
\end{table*}
% FB6522, FC8A46, FDA66F, FED9BF, FFF6EF
% \textbf{w/o WF} indicates the feedback model is not trained using  wrong feedback as in \cref{sec:both_correct_and_wrong}.

\subsection{\editeval: Unit-test-driven Feedback Evaluation}
\label{ssec:coffeeeval}

We present \textbf{\editeval} as our reliable reward function in \coffeegym. The key idea is to measure the helpfulness of feedback by gauging the correctness of the edited code produced by a small, but cheap editor model that properly aligns editing with feedback.
Specifically, given a problem description $q$, a wrong solution $\Tilde{y}$, and feedback $\hat{c}$ from a feedback model $\theta$, an editor model $\phi$ generates an edited code $y'$ by grounding on $\hat{c}$, \ie $y' = \phi(q, \Tilde{y}, \hat{c})$. The \editeval score is defined as the proportion of test cases for which the edited code $y'$ produces the expected output:
\begin{multline}
\text{\editeval}(q, \Tilde{y}, \hat{c}, \phi, \mathcal{T}) \\
= \frac{1}{k} \sum_{i=1}^{k} \mathds{1}\left(\phi(q, \Tilde{y}, \hat{c})(x_i) = z_i\right)
\end{multline}
where each element $t_i \in \mathcal{T}$ consists of an input $x_i$ and an expected output $z_i$, and $\mathds{1}$ is a binary indicator function that returns 1 if the output of $y'$ matches the expected output $z_i$. 
By reflecting the correctness of the edited code, the resulting score serves as an accurate measure for the effectiveness of the generated feedback in code editing.

\subsubsection{Training a Faithful Code Editor to Align Editing with Feedback}\label{sec:both_correct_and_wrong}

General code LLMs are trained to produce only correct codes, resulting in a bias toward correct editing regardless of feedback quality. To address this, we train a code editor $\phi$ that aligns its output with the helpfulness of the feedback by training the model to generate both correct edits $(q, y, c^*, y^*) \in \mathcal{D}_{correct}$ and incorrect edits $(q, y, \tilde{c}, \tilde{y}) \in \mathcal{D}_{wrong}$ in \cf. The training objective is defined as:
\begin{multline}
\mathcal{L}(\phi) = -\sum_{(q, y, c^*, y^*) \in \mathcal{D}_{correct}
} \log p_\phi( y^* \mid  q, y, c^*) \\
- \sum_{(q, y, \tilde{c}, \tilde{y}) \in \mathcal{D}_{wrong}}
\log p_\phi(\Tilde{y} \mid q, y, \tilde{c})
\end{multline}
To prevent confusion during training, we follow \citet{wang-etal-2023-scott} and indicate the correctness of the target code by prepending the keywords \texttt{[Correct]} and \texttt{[Wrong]} to the code sequence.

By learning from both positive and negative examples, the editor learns to conduct code editing by faithfully following the given feedback. It allows us to use the editor's output as a reliable metric for evaluating feedback generation models in our \coffeegym environment.

\section{Validating \editeval}

\subsection{Experimental Setting}
\paragraph{Implementation details.} We implement \editeval with DeepSeekCoder-7B model as the backbone in all our experiments.
For further details, please refer to Appendix~\ref{appendix:implementation_details}.

\subsection{Reliability of \editeval}
\label{sec:validate_eval}

\paragraph{Baselines.}
We compare our \editeval with two evaluation methods: G-Eval~\citep{liu2023geval} and Editing. 
For G-Eval, we directly assess feedback quality in Likert-scale (1 - 5) using score rubrics~\citep{kim2023prometheus}. 
Editing baselines follow the same evaluation scheme as \editeval but use general code LLMs for the editor $\phi$.
We consider with three code LLMs, GPT-3.5-Turbo, GPT-4-Turbo, and DeepSeek-Coder-7B.
The prompt we use for G-Eval is in Appendix~\ref{appendix:feedback_quality_eval}.

\paragraph{Evaluation.}
To measure the alignment between feedback generation and code editing, we use test set of \cfwemoji, where each $c$ is annotated with a binary label on its helpfulness.
For Editing methods (including ours), we regard the output as positive prediction when the edited code passes all test cases.
Also, we provide Pearson correlation coefficients for both Editing and G-Eval methods to analyze the correlation between the predicted score and the ground-truth labels.

\begin{table*}[t!]
\renewcommand{\arraystretch}{1.15} %1.15
\centering

\resizebox{\textwidth}{!}{
\begin{tabular}{lcccccccc}
\toprule
\multirow{2}{*}{\textbf{Methods}} & \multirow{2}{*}{\textbf{Params.}} & \multirow{2}{*}{\textbf{Open-source}} & \multicolumn{2}{c}{\textbf{HumanEvalFix}} & \multicolumn{2}{c}{\textsc{\textbf{Coffee-Test}}} & \multicolumn{2}{c}{\textbf{Average}} \\ \cmidrule(lr){4-5} \cmidrule(lr){6-7} \cmidrule(lr){8-9}
& & & \textbf{Pass@1} & \textbf{$\Delta$} & \textbf{Pass@1} & \textbf{$\Delta$} & \textbf{Pass@1} & \textbf{$\Delta$}  \\
\midrule
\cellcolor{gray!11}GPT-4-Turbo~\citep{openai2023gpt4} & \cellcolor{gray!11}- & \cellcolor{gray!11}\redcross & \cellcolor{gray!11}83.5 & \cellcolor{gray!11}- & \cellcolor{gray!11}43.8 & \cellcolor{gray!11}- & \cellcolor{gray!11}63.6 & \cellcolor{gray!11}- \\
\cellcolor{gray!11}GPT-3.5-Turbo~\citep{openai2023chatgpt} & \cellcolor{gray!11}- & \cellcolor{gray!11}\redcross & \cellcolor{gray!11}75.0 & \cellcolor{gray!11}- & \cellcolor{gray!11}32.2 & \cellcolor{gray!11}- & \cellcolor{gray!11}53.6 & \cellcolor{gray!11}- \\

\midrule
DeepSeek-Coder~\citep{Guo2024DeepSeekCoderWT} & 7B & \greencheck & 60.4 & - & 33.8 & - & 47.1 & - \\
~~+ Execution Feedback & - & \greencheck & 68.3 & + 7.9 & 38.3 & + 4.5 & 53.3 & + 6.2 \\
~~+ Self-Feedback & 7B & \greencheck & 67.7 & + 7.3  & 28.3 & - 5.5  & 48.0 & + 0.9  \\
~~+ OpenCodeInterpreter-DS-Coder Feedback & 7B & \greencheck & 64.6 & + 4.2 & 30.5 & - 3.3 & 47.5 & + 0.5  \\
~~\cellcolor{blue!10}+ \textbf{\textsc{Ours}} & \cellcolor{blue!10} \textbf{7B} & \cellcolor{blue!10}\greencheck & \cellcolor{blue!10}\textbf{73.8} & \cellcolor{blue!10}\textbf{+ 13.4} & \cellcolor{blue!10}\textbf{47.2} & \cellcolor{blue!10}\textbf{+ 13.4} & \cellcolor{blue!10}\textbf{60.5} & \cellcolor{blue!10}\textbf{+ 13.4} \\ \hdashline
~~\cellcolor{gray!11}+ GPT-3.5-Turbo Feedback & \cellcolor{gray!11}- & \cellcolor{gray!11}\redcross & \cellcolor{gray!11}72.5 & \cellcolor{gray!11}+ 12.1 & \cellcolor{gray!11}35.5 & \cellcolor{gray!11}+ 1.7 & \cellcolor{gray!11}54.0 & \cellcolor{gray!11}+ 6.9 \\
~~\cellcolor{gray!11}+ GPT-4-Turbo Feedback & \cellcolor{gray!11}- & \cellcolor{gray!11}\redcross & \cellcolor{gray!11}74.4 & \cellcolor{gray!11}+ 14.0 & \cellcolor{gray!11}44.4 & \cellcolor{gray!11}+ 10.6 & \cellcolor{gray!11}59.4 & \cellcolor{gray!11}+ 12.3 \\

\midrule
CodeGemma~\citep{codegemma_2024} & 7B & \greencheck & 53.7 & - & 14.4 & - & 34.1 & - \\
~~+ Execution Feedback  & - & \greencheck & \textbf{61.6} & \textbf{+ 7.9} & 15.0 & + 0.6 & 38.3 & + 4.2 \\
~~+ Self-Feedback & 7B & \greencheck & 53 & - 0.7  & 16.6 & + 2.2  & 34.8 & + 0.7  \\
~~+ OpenCodeInterpreter-DS-Coder Feedback & 7B & \greencheck & 36.5 & - 17.2 & 15 & + 0.6 & 25.8 & - 8.3 \\
~~\cellcolor{blue!10}+ \textbf{\textsc{Ours}} & \cellcolor{blue!10} \textbf{7B} & \cellcolor{blue!10}\greencheck & \cellcolor{blue!10}\underline{59.7} & \cellcolor{blue!10}\underline{+ 6.0} & \cellcolor{blue!10}\textbf{31.1} & \cellcolor{blue!10}\textbf{+ 16.7} & \cellcolor{blue!10}\textbf{45.4} & \cellcolor{blue!10}\textbf{+ 11.4} \\\hdashline
~~\cellcolor{gray!11}+ GPT-3.5-Turbo Feedback & \cellcolor{gray!11}- & \cellcolor{gray!11}\redcross & \cellcolor{gray!11}57.3 & \cellcolor{gray!11}+ 3.6 & \cellcolor{gray!11}22.2 & \cellcolor{gray!11}+ 7.8 & \cellcolor{gray!11}39.8 & \cellcolor{gray!11}+ 5.7 \\
~~\cellcolor{gray!11}+ GPT-4-Turbo Feedback & \cellcolor{gray!11}- & \cellcolor{gray!11}\redcross & \cellcolor{gray!11}65.8 & \cellcolor{gray!11}+ 12.1 & \cellcolor{gray!11}22.7 & \cellcolor{gray!11}+ 8.3 & \cellcolor{gray!11}44.3 & \cellcolor{gray!11}+ 10.2 \\

\midrule
OpenCodeInterpreter-DS-Coder~\citep{zheng2024opencodeinterpreter} & 7B & \greencheck & 65.8 & - & 30.5 & - & 48.1 & - \\
~~+ Execution Feedback  & - &  \greencheck & 66.4 & + 0.6 & 36.6 & + 6.1 & 51.5 & + 3.4 \\
~~+ Self-Feedback & 7B & \greencheck & 62.1 & - 3.7  & 21.1 & - 9.4  & 41.6 & - 6.5  \\
~~+ DeepSeek-Coder Feedback & 7B & \greencheck  & 56.1 & - 9.7 & 28.3 & - 2.2 & 42.2 & - 5.9 \\
~~\cellcolor{blue!10}+ \textbf{\textsc{Ours}} & \cellcolor{blue!10}\textbf{7B} & \cellcolor{blue!10}\greencheck & \cellcolor{blue!10}\textbf{70.1} & \cellcolor{blue!10}\textbf{+ 4.3} & \cellcolor{blue!10}\textbf{42.7} & \cellcolor{blue!10}\textbf{+ 12.2} & \cellcolor{blue!10}\textbf{56.4} & \cellcolor{blue!10}\textbf{+ 8.3} \\ \hdashline
~~\cellcolor{gray!11}+ GPT-3.5-Turbo Feedback & \cellcolor{gray!11}- & \cellcolor{gray!11}\redcross & \cellcolor{gray!11}68.3 & \cellcolor{gray!11}+ 2.5 & \cellcolor{gray!11}32.7 & \cellcolor{gray!11}+ 2.2 & \cellcolor{gray!11}50.5 & \cellcolor{gray!11}+ 2.4 \\ 
~~\cellcolor{gray!11}+ GPT-4-Turbo Feedback & \cellcolor{gray!11}- & \cellcolor{gray!11}\redcross & \cellcolor{gray!11}72.5 & \cellcolor{gray!11}+ 6.7 & \cellcolor{gray!11}43.3 & \cellcolor{gray!11}+ 12.8 & \cellcolor{gray!11}57.9 & \cellcolor{gray!11}+ 9.8 \\
\bottomrule

\end{tabular}
}
\caption{Code editing results of our feedback model trained with \coffeegym, \ie PPO-\editeval, on HumanEvalFix and \textsc{Coffee-Test}. We pair our feedback model with an open-source code LLM as the code editor.}
\label{tab:humanevalfix}
\end{table*}

\begin{figure}[!t]
\centering
    \includegraphics[width=1.0\linewidth]{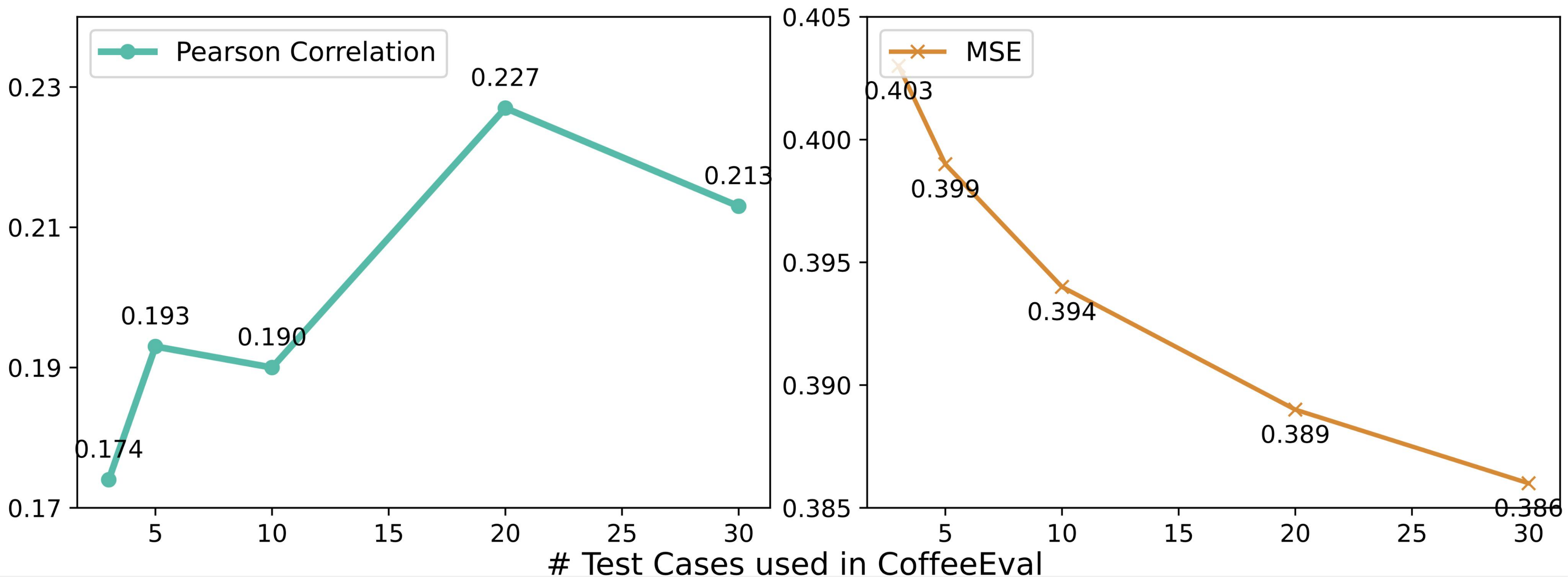}
\caption{Ablation results on the number of test cases used in \editeval. The evaluation performance decreases as the number of test cases declines.} 
\label{fig:test_case_analysis}
\end{figure}
\subsection{Results and Analysis}
\paragraph{\editeval faithfully aligns feedback quality with editing performance.}
As shown in Table~\ref{tab:evaluator_validation}, DeepSeek-\textsc{CoffeeEval} achieves higher Pearson correlation and lower MSE than all G-Eval and Editing baselines. In particular, our approach shows even higher correlation than the G-Eval baseline implemented with GPT-4-Turbo.
The strong performance of our \editeval validates its effectiveness in assessing the quality of NL feedback in the code editing task.

\paragraph{Code LLMs are skewed toward correct editing, regardless of the feedback quality.}
While code LLMs have shown promising results in code generation tasks, they do not faithfully reflect the helpfulness of feedback on code editing. 
Especially, GPT-4-Turbo, the current SOTA code LLM, shows the highest Pass@1 among baselines, but it also tends to generate correct code even with wrong feedback.
These results suggest that the training process with our pairwise feedback data is an essential step in building a reliable reward model.

\paragraph{The performance of \editeval benefits from the number of test cases.}
Figure~\ref{fig:test_case_analysis} compares the Pearson correlation coefficient and MSE with respect to the number of test cases.
We observe that a higher number of test cases leads to more accurate evaluation on the feedback quality, which validates our design choice of \coffeewemoji.

\section{Benchmarking Reference Methods of \coffeegym}
In this section, we apply the feedback model trained using \coffeegym on various open-source LLMs and assess its effectiveness in enhance code editing performance.
Furthermore, we comprehensively explore a wide range of training strategies available in our \coffeegym to provide insights on building helpful feedback models. 

\subsection{Effectiveness of \coffeegym in Training Feedback Models}

\subsubsection{Experimental Setting}\label{sec:end2end_val.compare_ref}
\paragraph{Implementation details.}
We train our feedback model based on DeepSeekCoder-7B using \coffeegym by applying PPO. Further details are in Appendix~\ref{appendix:coffeegym_details}.
\paragraph{Benchmarks.}
We test the feedback model trained using \coffeegym on HumanEvalFix~\citep{muennighoff2023octopack}, a widely used code editing benchmark. The task is to fix the errors in given erroneous code and the correctness of the edited code is measures by running the annotated test cases. Then, if the submitted solution passes all testcases the solution is evaluated as success and pass@1 is calculated as the percentage of the passed solutions for all promplems.
We carefully check if there is data leakage in \cf and verify there is no overlap between \cf and HumanEvalFix (Appendix~\ref{appendix:overlap}).
Additionally, we assess the effectiveness of our approach on a held-out test set named \textsc{Coffee-Test}. It consists of 180 instances of $(q, \Tilde{y}, y^*, \mathcal{T})$ pairs that are collected following the same process in \cref{ssec:coffee} but with no overlapping problems with \cf.\footnote{While we have considered other code editing benchmarks, DebugBench~\citep{tian2024debugbench} and CodeEditorBench~\citep{guo2024codeeditorbench}, we find that these benchmarks have a critical issue; even the ground-truth solution cannot pass the unit test. A detailed discussion on this issue is in Appendix~\ref{appendix:benchmarks}.}

\paragraph{Baselines.}
We compare with the following baselines that provides feedback for code editing: (1) Execution Feedback~\citep{chen2023teaching}: execution results of the generated code, \eg error messages, without using any LLMs , (2) Self-Feedback~\citep{madaan2023selfrefine}: NL feedback generated by the code editor itself, (3) OpenCodeInterpreter Feedback~\citep{zheng2024opencodeinterpreter}: a code LLM especially trained on Code-Feedback dataset. We also provide the results of feedback from closed-source LLMs, GPT-3.5-Turbo and GPT-4-Turbo, but these models are not our main focus as we aim to develop open-source feedback models.

\subsubsection{Results}
In Table~\ref{tab:humanevalfix}, we compare the performance of our best feedback model with other feedback methods using various open-source models. Consistent with the findings from \citet{chen2023teaching}, we observe improvements across all code LLMs when using Execution Feedback. However, we find that open-source code LLMs, despite their capabilities in the code domain, struggle to generate helpful NL feedback for code editing (Self-Feedback), highlighting the complexity of producing effective feedback. Notably, our approach demonstrates comparable performance to GPT-3.5/4-Turbo, significantly closing the performance gap between closed-source and open-source models in the task of feedback generation for code editing.

\subsection{Comparing Different Training Strategies in \coffeegym} \label{ssec:end2end_val.compare_feed}
\subsubsection{Experimental Setting}
\paragraph{Training strategies.}
For training algorithm, we explore DPO, PPO, and Rejection Sampling (RS).
In RS, we sample 10 $\hat{c}$ from SFT model, and collect $\hat{c}$ with top-1 \editeval score as labels for the next iteration of SFT. For PPO, we use \editeval as the reward model. We use 3 variants for DPO: (1) DPO-TS: We construct preference pair by selecting the teacher model's feedback (\ie GPT-3.5-Turbo) as $c^+$, and the student model's (SFT) response as $c^-$~\citep{tunstall2023zephyr}, (2) DPO-CW: We directly use the labeled feedback pair $(c^*, \Tilde{c})$. (3) DPO-\editeval: We sample 10 $\hat{c}$, same as RS, and we construct preference pair with $\hat{c}$ of top-1 and bottom-1 \editeval score.

\subsubsection{Results}
\begin{figure}[!t]
\centering

    \includegraphics[width=1.0\linewidth]{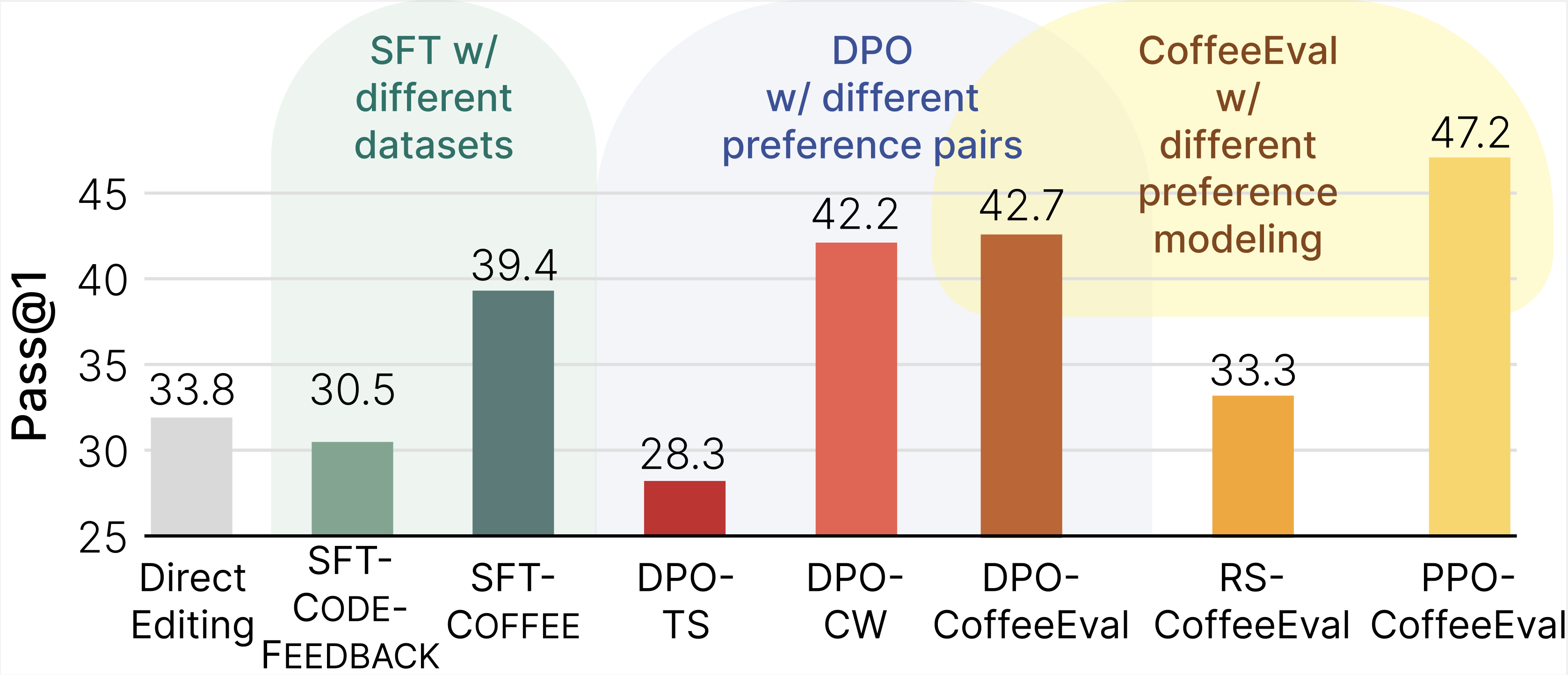}

\caption{End-to-end validation results of the reference methods in \coffeegym on \textsc{Coffee-Test}.}
\label{fig:coffeegym_validate}
\end{figure}

\paragraph{\cf provides helpful train data for SFT.}
In Figure~\ref{fig:coffeegym_validate}, we find that SFT-\cf provides more helpful feedback than SFT-\textsc{Code-Feedback} trained on Code-Feedback. This results suggest that \cf serves as a valuable resource for fine-tuning feedback models. 

\paragraph{\cf and \editeval allow informative preference pair construction for DPO.}
DPO-\editeval achieves the best results among DPO variants, closely followed by DPO-CW, which utilizes correct-wrong pairs from \cf.  However, DPO-TS significantly underperforms even with the correct feedback $c^+$ sampled from the teacher. 
We conjecture that the teacher's feedback may not always be superior to the student's, leading to suboptimal preference pairs.

\paragraph{PPO is the most effective training algorithm.}
PPO-\editeval outperforms DPO-\editeval and RS-\editeval, despite using the same reward model. We hypothesize that online RL methods like PPO allow for continuous updates on the reference model and lead to better alignment compared to offline methods like DPO, which learn from a fixed initial model.

\subsection{Analysis}
\begin{figure}[t]
\centering
    \includegraphics[width=1.0\linewidth]{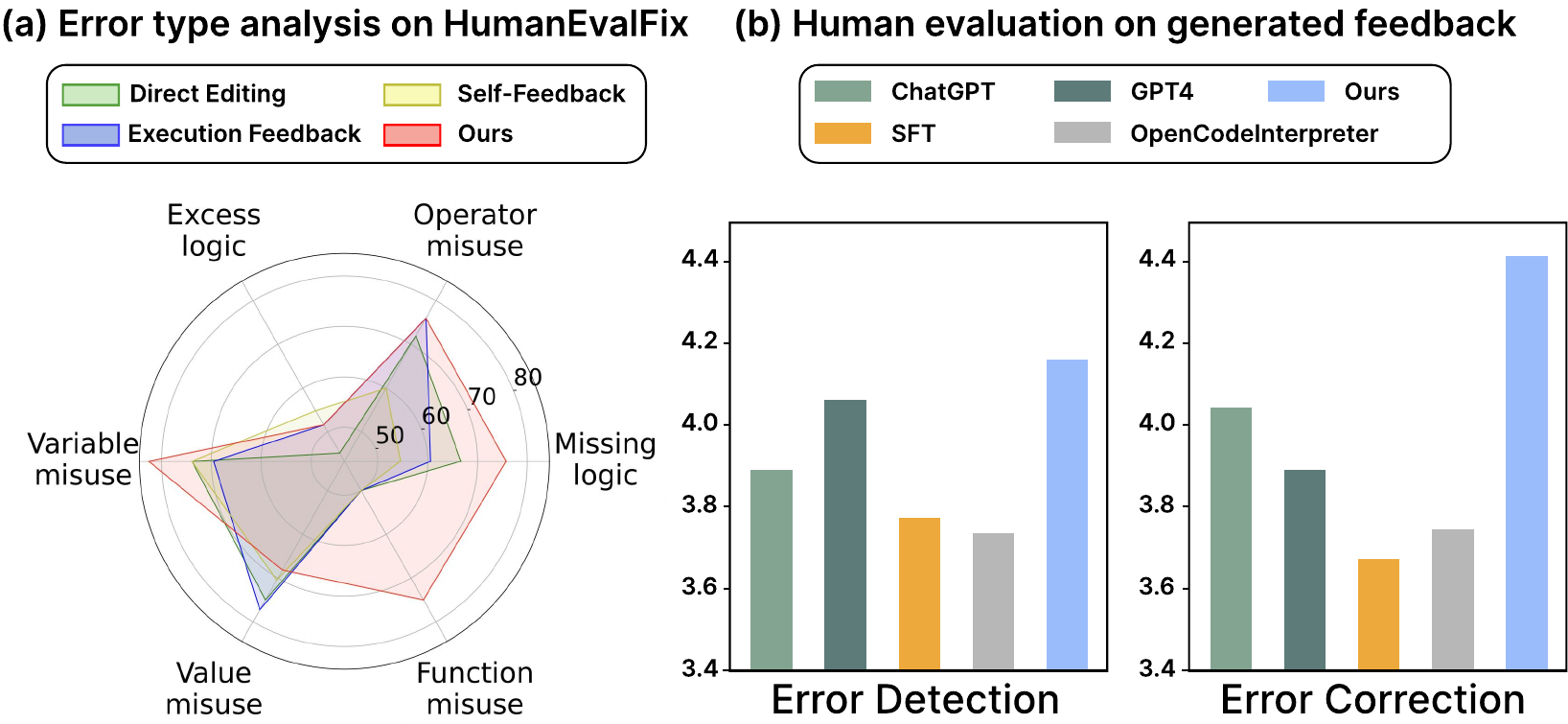}
\caption{(a) Breakdown of editing performance on HumanEvalFix by different error types. (b) Human evaluation of the feedback generated on HumanEvalFix. See Appendix~\ref{appendix:human_evaluation} for details on human evaluation.} 
\label{fig:error_analysis}
\end{figure}

\paragraph{Fine-grained analysis by error type.}
In Figure~\ref{fig:error_analysis}a, we compare the baselines with our approach across different error types. Our feedback model is particularly effective at correcting Missing logic and Function misuse errors, which can greatly benefit from NL feedback by providing a detailed explanation for editing. In value misuse, our model shows slightly lower performance. We posit that this is due to the discrepancy between the distribution of errors from human-authored data (\ie \cf) and synthetic data, where our model is tested. 
\paragraph{Human evaluation on feedback quality.}
To provide a more accurate analysis of the feedback quality, we conduct human evaluation using qualified workers from MTurk.\footnote{The details of our human evaluation are in  Appendix~\ref{appendix:human_evaluation}.} The results in Figure~\ref{fig:error_analysis}b show that the feedback from our model is rated as more helpful and informative compared to the baselines, supporting the findings in \cref{ssec:end2end_val.compare_feed}.

\section{Related Work}
\paragraph{Code editing.}

Code LLMs have shown promising code generation capabilities by training on massive code corpora~\citep{li2023starcoder, wang2023codet5plus}.
Despite their promising capabilities, there remains a possibility of errors, making code editing tasks essential for ensuring code quality and correctness~\citep{muennighoff2023octopack}.
In response to this necessity, recent studies have focused on assessing the code editing capabilities of code LLMs, by proposing new benchmarks for the task~\citep{tian2024debugbench,guo2024codeeditorbench}.

\paragraph{Refining with external feedback.}

In code editing, two types of widely used external feedback are execution feedback~\citep{gou2023critic, chen2023teaching} and NL feedback~\citep{madaan2023selfrefine, shinn2023reflexion}.
Recently, \citet{zheng2024opencodeinterpreter} explored both types of feedback and demonstrate that NL feedback outperforms execution feedback.
Concurrent to our work, \citet{ni2024next} explored building feedback model, but they do not provide the dataset used nor the model checkpoint.

\paragraph{RL in code generation tasks.}
A line of research has explored improving LLMs' code generation with RL by leveraging the unit test results as reward~\citep{le2022coderl, liu2023rltf, shen2023pangu}. 
While the design of \editeval is largely inspired by this line of work, we show that building reward model for feedback learning using unit test results is non-trivial, since code LLMs do not faithfully reflect feedback into editing (Table~\ref{tab:evaluator_validation}).

\section{Conclusion}
In this paper, we present a comprehensive study on building open-source feedback models for code editing. We introduce \coffeegym, an environment for training and evaluating feedback models, and share valuable insights from our experiments. We hope our work will encourage researchers to further explore feedback model development using \coffeegym and our findings, advancing the field of code editing with NL feedback.

\section*{Limitations}\label{sec:limitations}
% Programming language
\paragraph{Scope of editing.}
\coffeegym tackles the task of code editing with a particular focus on correcting errors in codes. 
This leaves room for improvement in our RL approach to consider the efficiency and readability of the edited codes.
Also, we mainly focus on editing incorrect source codes in a competitive programming setting. 
Some examples from our feedback model (Appendix~\ref{appendix:practical}) suggest that our approach can be further applied to practical programming problems, \eg those that involve machine learning libraries.
In future studies, \coffeegym can be further expanded to real-world software engineering settings with additional training on general code corpora~\citep{li2023starcoder}.

\paragraph{Using synthetic test cases for measuring reward.}
While running synthetic test cases and using the resulting pass rates might be a promising proxy for reward calculation, there might be edge cases where even erroneous codes pass the synthetic test cases. Further research can incorporate \citet{liu2023is} to make more challenging test cases that can rigorously identify erroneous codes.

\paragraph{Single programming language.}
Our implementation of \coffeegym is limited to a single programming language, \ie Python. However, future work might apply a similar strategy as ours to expand our model to a multilingual setting, where the model is capable of understanding and editing diverse programming languages such as Java.

\paragraph{Single parameter size and architecture.}
Lastly, we implement the feedback models only with one parameter size and architecture. However, future work can apply our method to models with larger parameter sizes (\eg DeepSeek-Coder 70B), which is expected to perform better in code editing. Our framework can also be further applied to other architectures, as our method is model-agnostic.

\section*{Ethical Considerations}
While our dataset originates from online competitive programming platforms, we have ensured the exclusion of personal information to maintain privacy standards.
Additionally, we are aware of the potential risks associated with texts generated by language models, which can contain harmful, biased, or offensive content.
However, based on our assessments, this risk is mostly mitigated in our work.
Lastly, there exists a risk of hallucination in the process of feedback generation and code editing, leading to incorrect edits. 
This emphasizes the need for careful application in our approach.

\section*{Acknowledgement}
This work was supported by Institute of Information \& Communications Technology Planning \& Evaluation (IITP) grant funded by the Korean government (MSIT)(No.RS-2020-II201361, Artificial Intelligence Graduate School Program (Yonsei University)) and (No.RS-2021-II212068, Artificial Intelligence Innovation Hub) and (2022-0-00077, RS-2022-II220077,AI Technology Development for Commonsense Extraction, Reasoning, and Inference from Heterogeneous Data). Jinyoung Yeo is the corresponding author.

\bibliography{anthology, custom}

\appendix
% \section*{Appendix}
\appendix
\clearpage
\renewcommand{\thesection}{\Alph{section}}

\section{Details of \coffeegym}

\subsection{Details of \coffee~\textsc{Coffee}}
\label{appendix:dataset}

\subsubsection{Feedback Annotation}
\label{appendix:feedback_annotation}
We annotate both correct and wrong feedback for our dataset using GPT-3.5-Turbo.
We apply top-$p$ sampling and temperature, where $p = 0.95$ and $T = 0.7$. 
We limit the number of generation tokens to 500. 
We leave out submission histories where the LLM fails to find any errors. 
We also filter out submissions from different users whose correct solutions are identical, as these solutions are usually copied from the web without undergoing editing processes. 
With collected user's submission history $\{\Tilde{y}_1, \Tilde{y}_2, ..., y^*_n\}$, we sample correct edit pairs $\{\Tilde{y}_k, y^*_n\}_{k=1}^{n-1}$ to annotate correct feedback.
To annotate the wrong feedback, we use sequential pairs $\{\Tilde{y}_k, \Tilde{y}_{k+1}\}_{k=1}^{n-2}$ to capture transitions between consecutive incorrect solutions.
% and user's wrong edit traces $\{\Tilde{y}_k, \Tilde{y}_{k+1}\}_{k=1}^{n-2}$ to annotate wrong feedback.
The prompts used for annotating correct and wrong feedback are demonstrated in Appendix~\ref{appendix:correct_feedback_prompt} and Appendix~\ref{appendix:wrong_feedback_prompt}.

\subsubsection{Quality Analysis on Annotated Feedback}
To thoroughly analyze the quality of the feedback from GPT-3.5-Turbo, we conduct a human evaluation.
We ask human raters from Amazon Mechanical Turk (AMT) to score the quality of the feedback on a Likert scale.
To ensure proficiency, we filter out human raters who have not passed our qualification test, which assesses their knowledge of programming languages, especially Python.
From the test set of \textsc{Coffee}, we sample 100 instances for the evaluation.

On average, the annotated feedback is scored 3.88 with 0.91 STD, which suggests that the quality of the annotated feedback is generally acceptable by humans. 
The full distribution of the evaluation results is shown in Table~\ref{tab:correctness_score}.

\subsubsection{Synthesizing Test Cases}
\label{appendix:test_case}
We prompt GPT-3.5-Turbo to synthesize input test cases given a problem description with three demonstrations. 
For each test case, we execute the correct code to obtain the corresponding output. If execution was successful, we then pair these inputs and outputs to create sample input-output pairs. On average, we synthesize 35 test cases per problem.
We provide the prompt for the test case generation in Appendix~\ref{appendix:test_case_prompt}.

\subsubsection{Analysis on Machine-generated Test Cases}
To gain insights into the effectiveness of our machine-generated test cases, we conduct analyses exploring two key aspects: validity and diversity.

\begin{table}[t]
\small
\renewcommand{\arraystretch}{1.0}
\centering

\begin{tabular}{lc}
    \toprule
    \textbf{Correctness Score} & \textbf{Frequency (\%)} \\
    \midrule
    1  &  2 (0.6\%) \\
    2  & 21 (7.0\%) \\
    3  & 70 (23.3\%) \\
    4  & 126 (42.0\%) \\
    5  & 81 (27.0\%) \\
    \bottomrule
\end{tabular}

\caption{Distribution of human evaluation scores for GPT-3.5-Turbo feedback quality.}
\label{tab:correctness_score}
\end{table}

\paragraph{Validity of test cases.}
A critical question in evaluating our test suite is whether any incorrect solutions manage to pass all the test cases. 
To address this, we conducted an experiment using the evaluation set of the \textsc{Coffee} dataset. 
We randomly sampled 200 wrong code instances and calculated the pass ratios of the wrong codes. 
We show the statistics of the distribution of pass ratios.

\begin{table}[t]
% \small
% \renewcommand{\arraystretch}{1.15} %1.15
\centering

\resizebox{\columnwidth}{!}{
    \begin{tabular}{lccccccc}
    \toprule
    \textbf{}&\textbf{mean}&\textbf{std}&\textbf{min}&\textbf{25\%}&\textbf{50\%}&\textbf{75\%}&\textbf{max}\\
    \midrule
    Pass ratio & 0.342 & 0.370 & 0.000 & 0.000 & 0.162 & 0.693 & 0.985 \\
    
    \bottomrule
    \end{tabular}  
}

\caption{Pass ratio for incorrect code samples in the evaluation set of \textsc{Coffee} dataset.}
% The pass ratio is defined as the number of test cases passed divided by the total number of test cases.}
\label{tab:validity}
\end{table}

As shown in Table~\ref{tab:validity}, the maximum pass ratio is 0.985, which suggests that there are no wrong solutions that passed all the test cases. 
The mean score is 0.342, indicating that on average, wrong solutions fail the majority of the test cases. 
We further analyze the \textsc{Coffee-Test} and verified that no wrong solutions pass all the test cases.

\paragraph{Diverse difficulty of test cases.}
To demonstrate that our generated test cases cover a range of difficulties, we analyzed the pass ratio distribution for incorrect code samples annotated in the dataset.
We focused on a single problem from the \textsc{Coffee} evaluation set.

\begin{figure}[t]
\centering
    \includegraphics[width=1\linewidth]{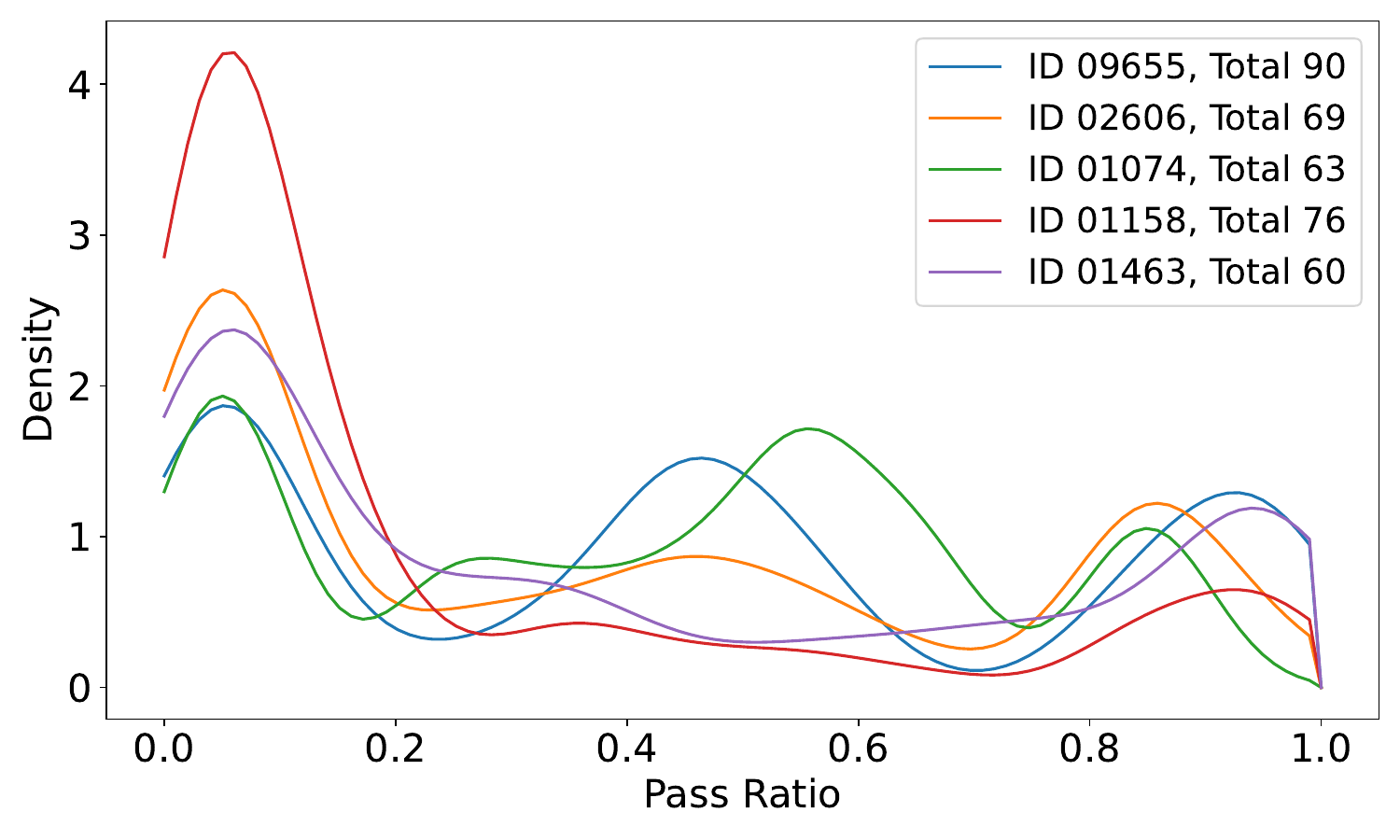}
\caption{Kernel Density Estimation plot of the pass ratio distribution for incorrect code samples.}
\label{tab:diverse_diff}
\end{figure}

As shown in Figure~\ref{tab:diverse_diff}, the results revealed that various incorrect solutions for this problem exhibited different pass ratios, indicating that our test cases encompass diverse difficulty levels.

\subsubsection{Data Analysis}
\label{appendix:error_code_analysis}
We conduct following experiments to explore original features in \cf dataset.
\paragraph{Length of edit trace}
We analyze the distribution of average length of edit trace by problem level. In Figure~\ref{fig:violin}.a, we observe a steady increase in the average length of edit traces from human programmers with increasing difficulty levels. This suggests that problems in \cf are challenging for human programmers, as they tend to make more incorrect submissions for problems with higher difficulty levels.  
%Consequently, we effectively obtain user's edit traces even for high-difficulty problems that are challenging for the model to solve or generate plausible edit traces for.
\paragraph{Code diversity.}
To assess the diversity of human-written codes compared to machine-generated codes, we conduct a similarity analysis on error codes. 
Specifically, we sample problems from \cf where more than 100 users submitted solutions and collect the wrong code from these users. 
We also sample an equal number of wrong codes from ChatGPT and GPT-4 with top-p sampling of $p = 0.95$ and temperature $T = 0.6$. 
For each set of incorrect solutions sampled from user solutions, ChatGPT, and GPT-4, we use CodeBERT~\citep{feng2020codebert} to compute embeddings for incorrect solutions and measure cosine similarity for all possible pairs in the set.

Figure~\ref{fig:violin}.b shows the histogram of the number of problems by the average embedding similarity of incorrect solution pairs. 
We find that machine-generated codes (\ie ChatGPT, GPT4) tend to be more similar to each other than human-generated codes, indicating that collecting human-generated code allows for more diverse set of wrong code samples. 
\paragraph{Code complexity} To show that problems in \cf are challenging for code LLMs, we measure the code generation performance of GPT-4 using Pass@1 and compare it with the solve rate of human programmers. Note that the latter is given as the metadata from the programming platform and computed as the proportion of correct solutions among all solutions submitted for problems in \cf.
The results (Figure~\ref{fig:violin}.c) suggest that even the state-of-the-art LLM, \ie GPT-4, struggles to produce correct solutions for problems in \cf and lags behind human programmers.

\subsubsection{Analysis on Train-test Overlap}\label{appendix:overlap}
\begin{figure}[t!]%1.15
    \centering
        \begin{subfigure}[b]{0.7\linewidth}
        \includegraphics[width=\textwidth]{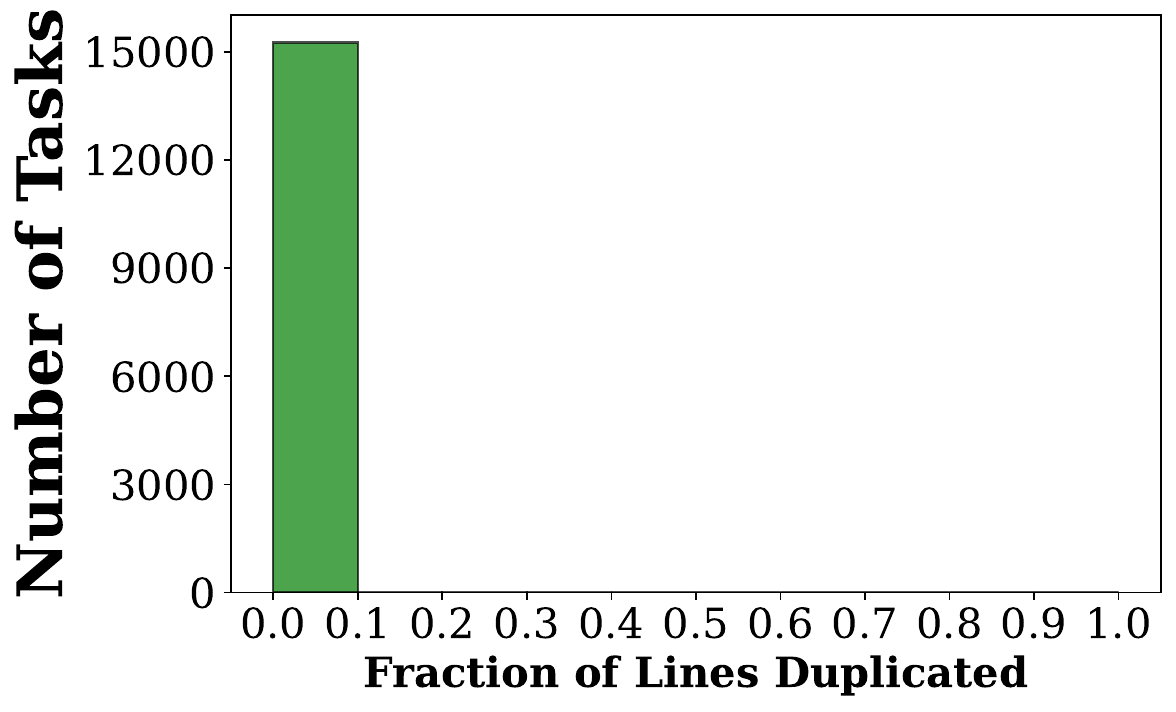}    
        \caption{Fraction of line overlaps.}
        \end{subfigure}
        \begin{subfigure}[b]{0.7\linewidth}
        \includegraphics[width=\textwidth]{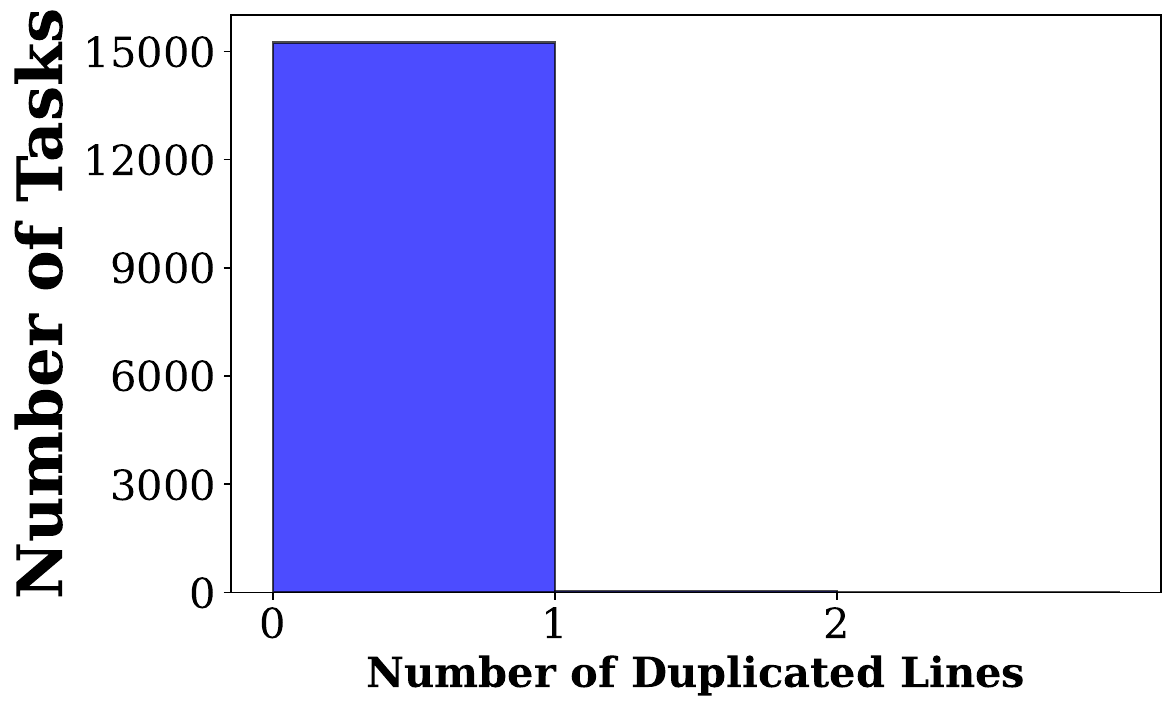}
        \caption{Absolute number of line overlaps.}    
        % \caption{Quality evaluation of feedback by ChatGPT}   
        \end{subfigure}
    \caption{Analysis on train-test overlap between \cf and HumanEval.}
\label{fig:overlap-humaneval}
\end{figure}

A possible concern is that the training data in \textsc{Coffee} might overlap with the test data in the code benchmark (\ie HumanEval). Therefore, we follow \citet{augustus2021program} and measure the amount of identical codes (based on the number of repeated lines) between the training and test data.
Figure~\ref{fig:overlap-humaneval} reports both the fraction and the absolution number of line overlaps between \cf~and HumanEval.
We observe that most solutions in \cf~do not contain lines that appear in the benchmark dataset which we evaluate our models on.

\subsection{Details of \editeval}

\subsubsection{Implementation Details}
\label{appendix:implementation_details}
We use DeepSeekCoder-7b\footnote{\url{https://huggingface.co/deepseek-ai/deepseek-coder-6.7b-instruct}}  as our backbone model using QLoRA~\citep{dettmers2023qlora}, incorporating 4-bit quantization with a learning rate of 5e-5 and a batch size of 4 for 2 epochs.
The training is run on 8 NVIDIA GeForce RTX 3090 GPUs. 
Regarding the LoRA configuration, we specify the dimension of low-rank matrices as 64, and alpha as 16.

\subsubsection{Training Details}
Following the approach of \citet{wang-etal-2023-scott}, we train the editor in two phases.
The initial phase includes the keywords \texttt{[Correct]} and \texttt{[Wrong]} in the code sequence, while the second phase trains the model without these keywords.
\paragraph{Phase I.}
We finetune our editor model $\phi$ using pairwise data of correct edits $(q, y, c^*, y^*) \in \mathcal{D}_{correct}$ and incorrect edits $(q, y, \tilde{c}, \tilde{y}) \in \mathcal{D}_{wrong}$ in \cf. 
During this phase, we additionally append keyword tokens $t^*$ and $\tilde{t}$ (\texttt{[Correct]} and \texttt{[Wrong]} respectively) with the target code sequences $y^*$ and $\tilde{y}$.
Therefore, the training objective for the initial phase is defined as:
\begin{multline}
\mathcal{L}(\phi) = \\
-\sum_{(q, y, c^*, y^*) \in \mathcal{D}_{correct}
} \log p_\phi(t^*, y^* \mid  q, y, c^*) \\
- \sum_{(q, y, \tilde{c}, \tilde{y}) \in \mathcal{D}_{wrong}}
\log p_\phi(\Tilde{t}, \Tilde{y} \mid q, y, \tilde{c})
\end{multline}

\paragraph{Phase II.}
After training the editor in Phase I, we continually train the editor model using the same dataset but without the keyword tokens.
Thereby, the training object for Phase II is defined as:
\begin{multline}
\mathcal{L}(\phi) = -\sum_{(q, y, c^*, y^*) \in \mathcal{D}_{correct}
} \log p_\phi( y^* \mid  q, y, c^*) \\
- \sum_{(q, y, \tilde{c}, \tilde{y}) \in \mathcal{D}_{wrong}}
\log p_\phi(\Tilde{y} \mid q, y, \tilde{c})
\end{multline}
We used the same hyperparameter settings in both phases and the prompt for training the code editor in Appendix~\ref{appendix:code_editor_prompt},

\subsection{Details of Reference Methods in \coffeegym}
\label{appendix:coffeegym_details}
\paragraph{Preference Tuning.}
Given a problem description, a wrong code, and the corresponding preference set, we apply Direct Preference Optimization (DPO)~\citep{rafailov2023direct} to train our critic. That is, we tune critic model to be biased towards helpful feedback. 

\paragraph{PPO.}
PPO optimizes the following objective:
\begin{multline}
\mathcal{L}_{\text{PPO}}(\theta) = \\
\hat{\mathbb{E}}_t \left[\min\left(r_t(\theta) \hat{A}_t, \text{clip}(r_t(\theta), 1 - \epsilon, 1 + \epsilon) \hat{A}_t\right)\right]    
\end{multline}
where $r_t(\theta)$ is the probability ratio between the current policy $\theta$ and the old policy $\theta_{\text{old}}$, $\hat{A}_t$ is an estimator of the advantage function at timestep $t$, and $\epsilon$ is a hyperparameter that controls the clipping range.

\paragraph{DPO.} From SFT model we sample 10 feedback strings and score them with \editeval. Among the 10 feedback collect feedback with top-1 score and bottom-1 score and construct preference pair, \ie $(c^+, c^-)$, for DPO training. Using this dataset, we additionally conduct DPO training on SFT model.

\paragraph{Rejection sampling.} From SFT model we sample 10 feedback strings and score them with \editeval. Among the 10 feedback we only collect feedback with top-1 score and construct dataset for further training. Using this dataset, we additionally conduct SFT.

\paragraph{Terms and License.} For our implementation and evaluation, we use Huggingface, TRL and vLLM library.\footnote{\url{https://huggingface.co/}} Both libraries are licensed under Apache License, Version 2.0.
We have confirmed that all of the artifacts used in this paper are available for non-commercial scientific use.

\section{Experimental Details}\label{appendix:experimental_details}
% \subsection{Baselines}\label{appendix:baselines}
% For our experiments, we consider the following open-source baselines:

% \paragraph{Code {Llama}.}
% Code Llama~\citep{roziere2023code} refers to variants of LLaMA2, specialized in code domains via fine-tuning on code corpus. 
% This collection includes various models tailored for specific uses: the foundation model (Code Llama), a Python-focused model (Code Llama-Python), and an instruction-following model (Code Llama-Instruct). 
% These models are available in sizes of 7B, 13B, and 34B parameters. 
% In our experiments, we use the Code Llama-Instruct model as the baseline.

% For closed-source baselines we consider ChatGPT~\citep{openai2023chatgpt} and GPT-4~\citep{openai2023gpt4}. Note that GPT-4 has shown the SOTA performance in code-related tasks. As these models are sensitive to input prompt, we use the prompts used in \citet{muennighoff2023octopack} to evaluate thes models.

\subsection{Benchmarks}
\label{appendix:benchmarks}

For our experiments, we consider the following benchmarks:
\paragraph{HumanEvalFix}
HumanEvalFix is a task of HumanEvalPack, manually curated using solutions from HumanEval~\citep{chen2021codex} for the task of code editing.
Given an (i) incorrect code function, which contains a subtle bug, and (ii) several unit tests (\ie test cases), the model is tasked to correct/fix the function. 
The dataset consists of 164 samples from the HumanEval solutions, and each sample comes with human-authored bugs across six different programming languages, thus covering 984 bugs in total. 
The bugs are designed in a way that the code is executed without critical failure but fails to produce the correct output for at least one test case.

We have confirmed that the dataset is licensed under the MIT License and made available for non-commercial, scientific use.
\paragraph{Reason for exclusion.}
We excluded DebugBench and CodeEditorBench for the following reasons:
\begin{itemize}
\item\textbf{DebugBench}~\citep{tian2024debugbench} is a debugging benchmark consisting of 4253 instances with 4 major categories and 18 minor types of bugs.
The metric is based on the test suites provided by LeetCode, requiring API calls for evaluation.
Due to the huge amount of API calls, LeetCode blocked the access during the evaluation, which lacked the accurate scoring.
Also, some questions were graded incorrectly even though ground-truth solutions were given.
Therefore, we decided not to use DebugBench for evaluation.
\item\textbf{CodeEditorBench}~\citep{guo2024codeeditorbench} is the framework designed for evaluating the performance of code editing.
Code editing is categorized into four scenarios, debugging, translation, polishing, and requirement switching, where our main focus is on debugging.
Similar to DebugBench, ground-truth solutions could not pass the unit test for some questions.
Also, functions imported from external python files and some specific packages were used in questions without details, which made the question imprecise.
So, we sent CodeEditorBench out of our scope.
\end{itemize}
\subsection{Metrics}
We use Pass@1 score to measure the code editing performance for all benchmarks. Specifically, Pass@1 is computed as the expected value of the correct rate per problem, when $n$ samples were generated to count the number of correct samples $c$ for each problem.
\begin{equation}
    \text{Pass@1} = \underset{\text{Problems}}{\mathbb{E}}\left[\frac{c}{n}\right] \times 100
\end{equation}

% \subsection{Analysis on Train-test Overlap}\label{appendix:overlap}
% A possible concern is that the training data in \textsc{Coffee} might overlap with the test data in the code benchmark (\ie HumanEval). Therefore, we follow \citet{augustus2021program} and measure the amount of identical codes (based on the number of repeated lines) between the training and test data.
% Figure~\ref{fig:overlap-humaneval} reports both the fraction and the absolution number of line overlaps between \cf~and HumanEval.
% We observe that most solutions in \cf~do not contain lines that appear in the benchmark dataset which we evaluate our models on.

\subsection{Feedback Quality Evaluation}
\label{appendix:feedback_quality_eval}
To assess the feedback quality in Likert-scale, we use G-Eval~\citep{liu2023geval} and prompt GPT-4-Turbo to evaluate the feedback quality.
Specifically, given problem description, input and output format, wrong code, and the corresponding feedback, we prompt GPT-4 to classify the feedback into one of the following five categories.
\begin{itemize}
    \item \textbf{Completely incorrect}: Feedback has no valid points and is entirely misleading.
    \item \textbf{Mostly incorrect}: Feedback has some valid points but is largely incorrect or misleading.
    \item \textbf{Neutral or somewhat accurate}: Feedback is partially correct but contains significant inaccuracies or omissions.
    \item \textbf{Mostly correct}: Feedback is largely accurate with only minor mistakes or omissions.
    \item \textbf{Completely correct}: Feedback is entirely accurate and provides a correct assessment of the code.
\end{itemize}
We apply the same top-$p$ sampling and temperature in Table~\ref{appendix:feedback_annotation} and include the prompt used for the evaluation in Appendix~\ref{appendix:g_eval}.

\subsection{Human Evaluation on Quality of Feedback}\label{appendix:human_evaluation}

\paragraph{Task description.}
The error detection and correction scores were determined by human annotators evaluating feedback on incorrect code using a Likert scale. 
The error detection score evaluates how accurately the feedback identifies errors in the incorrect code, while the error correction score assesses the correctness and effectiveness of the corrections suggested in the feedback.

\paragraph{Preparing feedback for the evaluation.}
We aim to analyze the quality of the feedback generated for code editing. We randomly sample 100 codes from \textsc{Coffee-Test} to assure the correctness of our evaluation.
For generating feedbacks, we use the erroneous codes provided in the dataset.

\begin{figure*}[t]
\centering
    \includegraphics[width=0.92\linewidth]{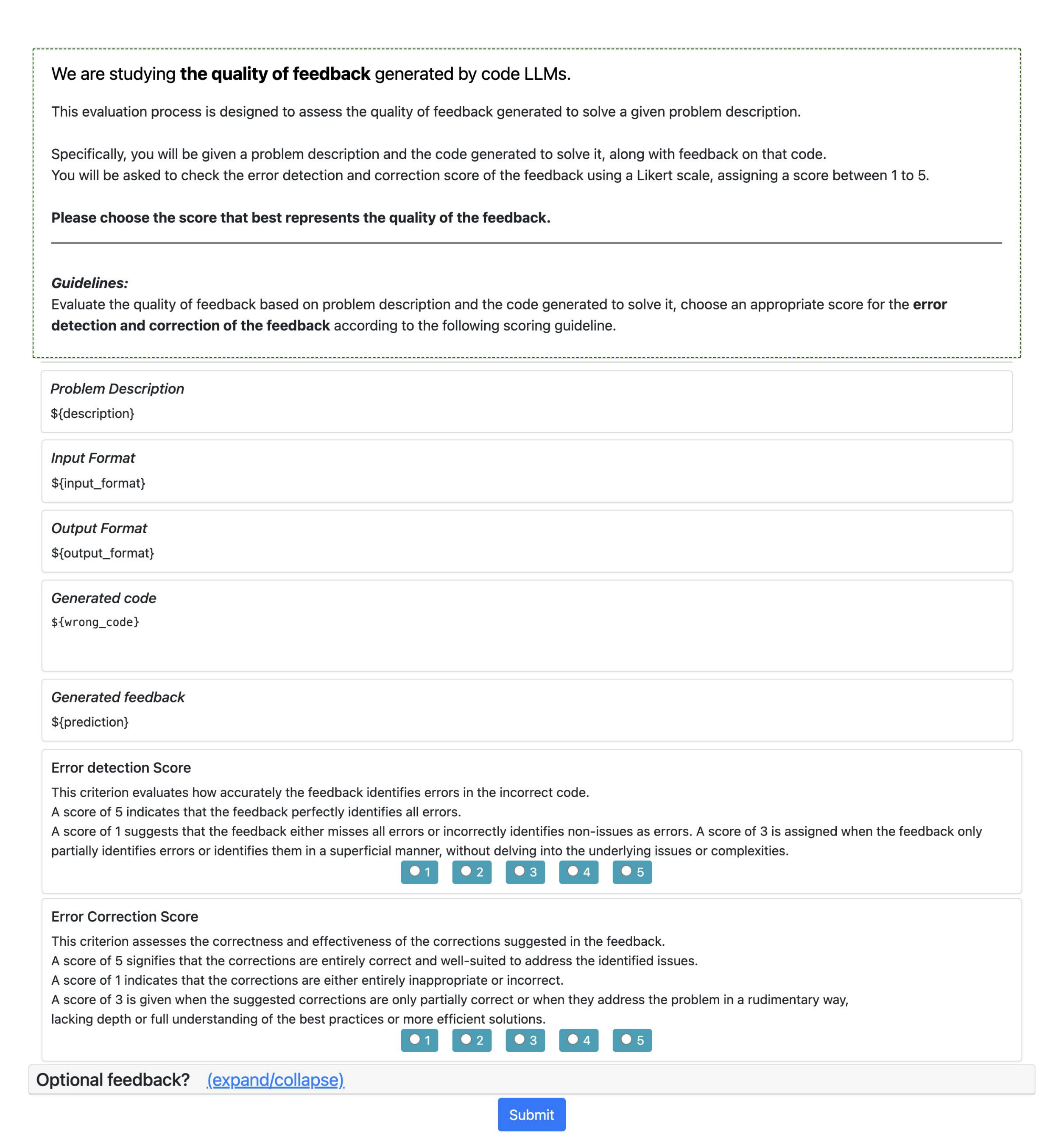}
\caption{The interface used for human evaluation on the feedback.} 
\label{fig:amt}
\end{figure*}

\paragraph{Details on human evaluation.}
We conduct human evaluation by using Amazon Mechanical Turk (AMT), which is a popular crowd sourcing platform.
As we need workers who have enough experience with Python, we conduct a qualification test to collect a pool of qualified workers.
In result, we recruit 186 workers who have passed the test, and task them to evaluate the quality of the feedback on Likert scale, ranging from 1 to 5.
Each sample is evaluated by three different raters to ensure the reliability.
Based on our estimates of time required per task, we ensure that the effective pay rate is at least \$15 per hour.
We use the evaluation interface in Figure~\ref{fig:amt}.
% We also report the Fleiss kappa~\citep{fleiss1971measuring} in Table~\ref{tab:fleiss_kappa} to evaluate the consistency of the evaluation among annotators.

\begin{figure}[t]
\centering
    \includegraphics[width=0.9\linewidth]{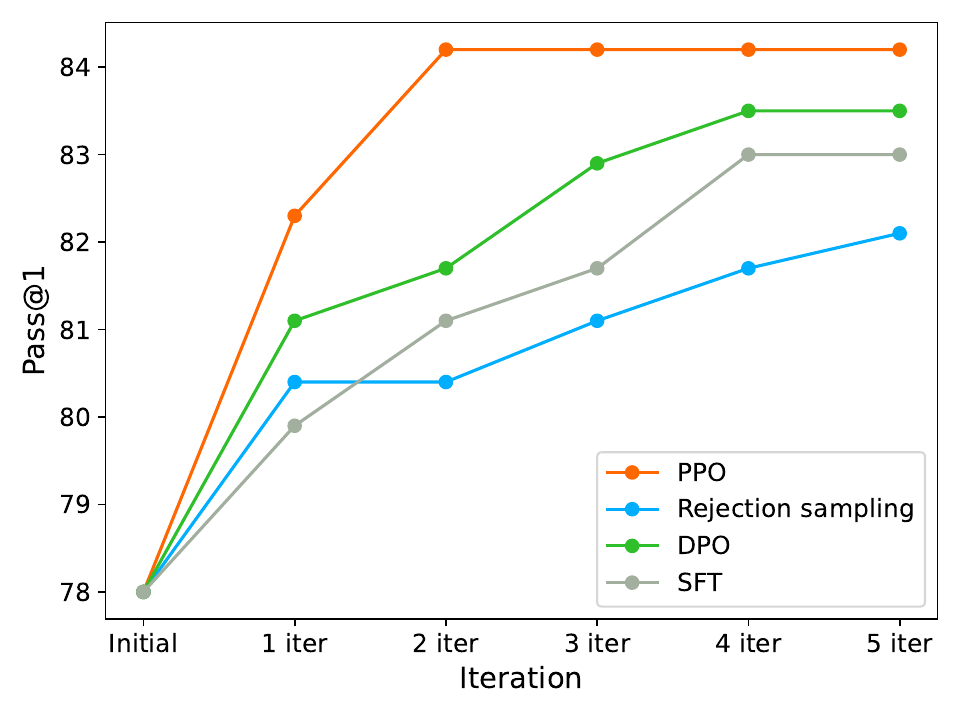}
\caption{Performance on test cases from HumanEval, measured under the iterative edit setting.}
\label{fig:iteration}
\end{figure}

\section{Additional Analysis}

\subsection{Iterative Editing}\label{appendix:iteration}
Inspired by \citet{zheng2024opencodeinterpreter}, we consider a practical setting where models are tasked with iterative code generation with feedback. 
We employed OpenCoderInterpreter-DS-7b as our codeLLM and used our feedback model to provide evaluations on the generated code. 
Our experiments included comparisons with reference methods in \coffeegym.
As shown in Figure~\ref{fig:iteration}, using our feedback model consistently enhanced performance over successive iterations. Consistent with our main experiment findings, both PPO and DPO improved feedback quality more effectively than rejection sampling. These results underscore the practical applications of our approach.

\subsection{Practical Programming Problems}
\label{appendix:practical}
To further explore the applicability of our feedback model (PPO-\editeval) to practical programming problems and assess its robustness across different domains, we conducted experiments using NumpyEval~\citep{CERT}. 
This dataset focuses on the general coding domain, specifically involving problems related to the NumPy library. We chose this benchmark to test our model's performance on unseen domains and evaluate its generalizability beyond our initial scope.
We utilized OpenCodeInterpreter-DS-Coder-7b as both the generation and editing model, while PPO-CoffeeEval served as the feedback model. To establish a baseline, we compared our approach against a Self-Feedback method, which used OpenCodeInterpreter-DS-Coder-7b for feedback as well.

As shown in Table~\ref{tab:numpy_eval_score}, our PPO-CoffeeEval model outperforms the baseline. 
These results suggest that our feedback model is not overfitted to Coffee dataset, and did not lost generalization ability to unseen domains.

\begin{figure*}[ht]
\centering
    \includegraphics[width=1.3\columnwidth]{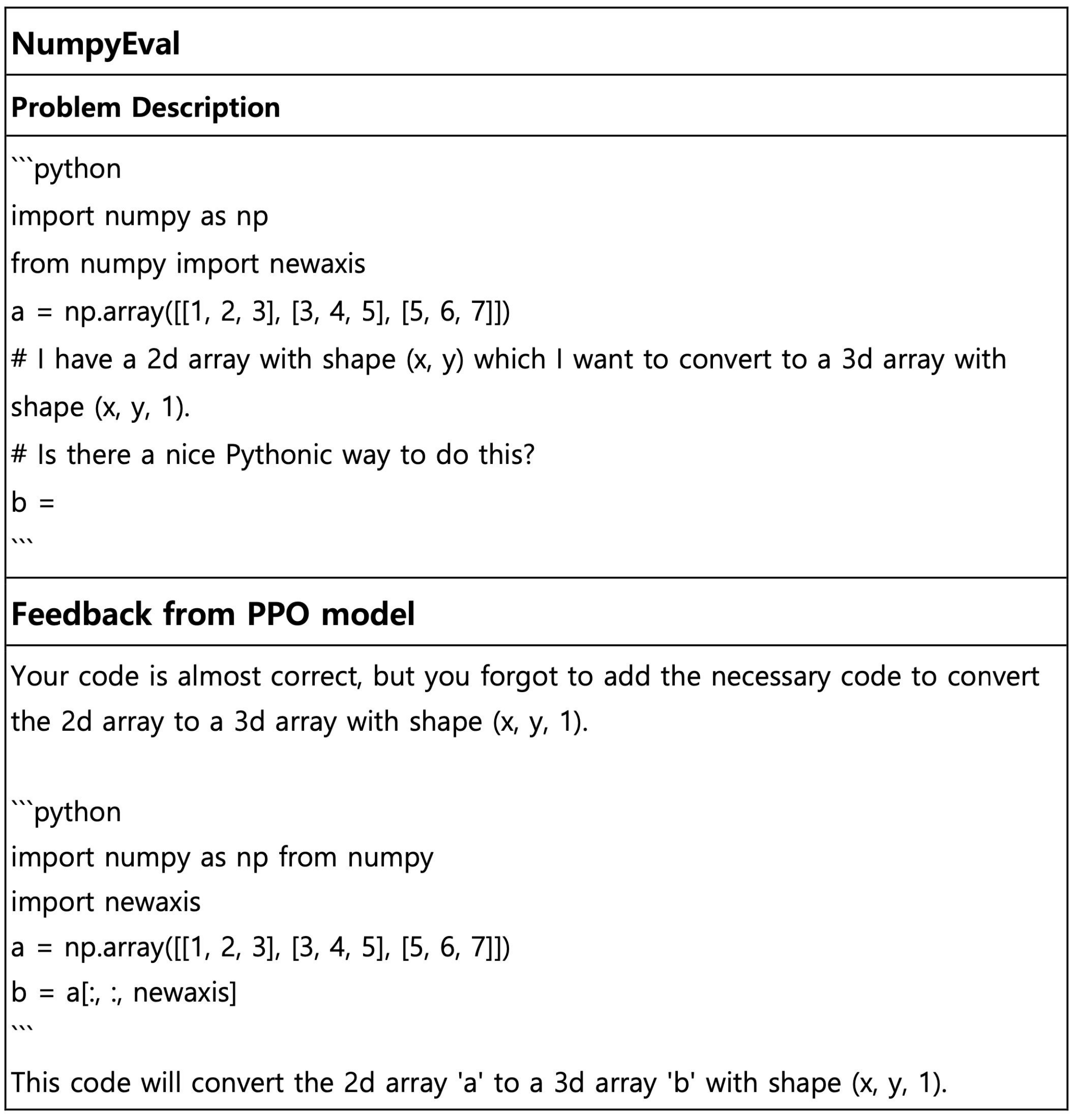}
\caption{Examples of the feedback from the PPO model on NumpyEval.}
\label{fig:case_study_numpyeval}
\end{figure*}

\begin{figure*}[ht]
\centering
    \includegraphics[width=1.3\columnwidth]{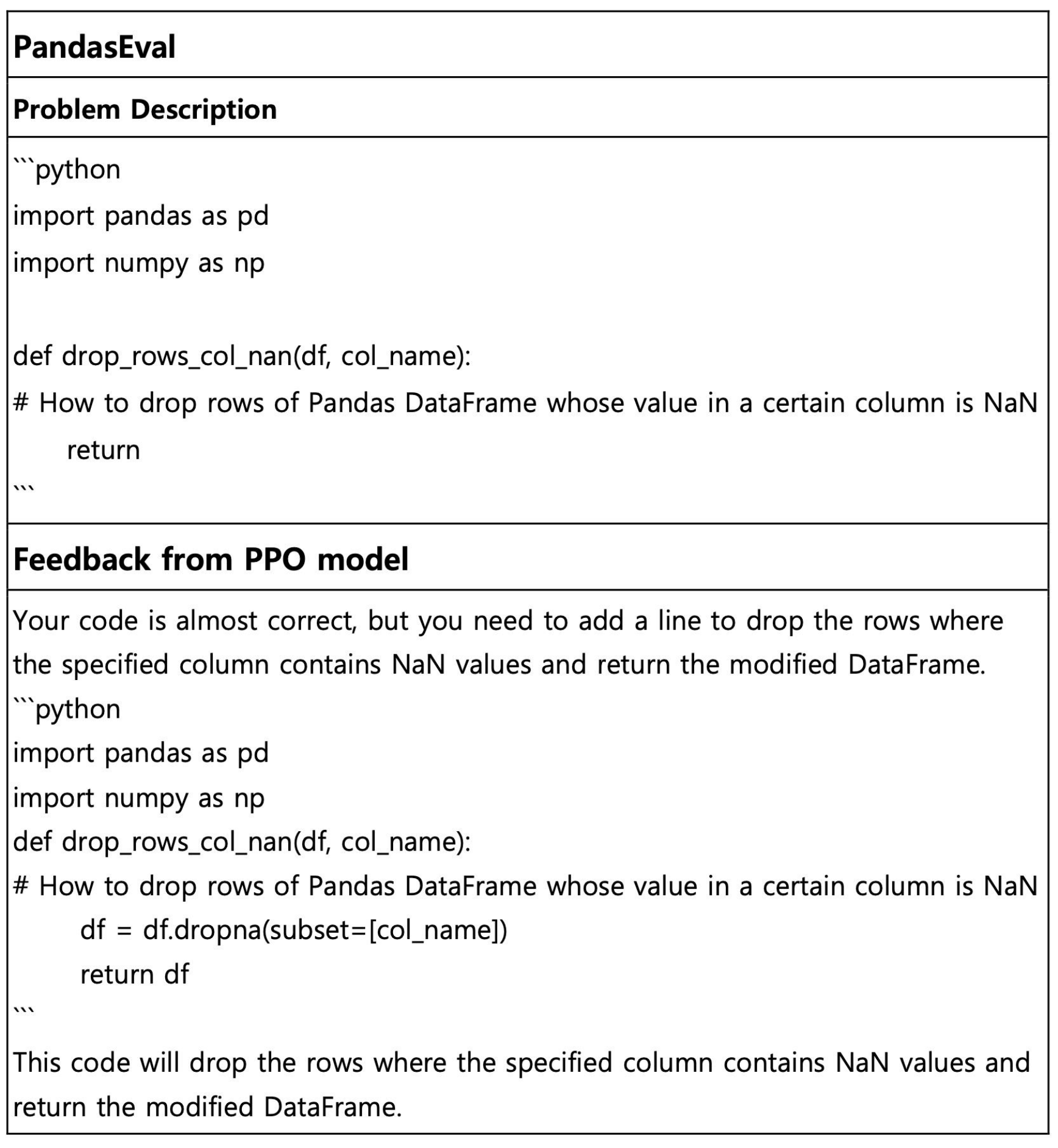}
\caption{Examples of the feedback from the PPO model on PandasEval.}
\label{fig:case_study_pandaseval}
\end{figure*}

For further analysis, we conducted a case study to examine the model's performance in more detail.
As illustrated in Figure~\ref{fig:case_study_numpyeval} and Figure~\ref{fig:case_study_pandaseval}, our model demonstrates the ability to generate helpful feedback even when the problem description is provided in Python comments rather than natural language format. 
In some instances, the feedback includes the necessary editing code.
This capability highlights the potential for using our model in practical scenarios, where users' queries can take various forms and formats, enhancing its applicability in real-world programming environments.

\begin{table}[t]
\small
\centering

\resizebox{\columnwidth}{!}{
    \begin{tabular}{lc}
    \toprule
    \textbf{Model}&\textbf{Pass@1}\\
    
    \midrule
    OpenCodeInterpreter-DS-Coder-7b  &  68.3 \\
    + PPO-\textsc{CoffeeEval}  & 70.3 \\
    
    \bottomrule
    \end{tabular}  
}

\caption{The performance of different feedback models on NumpyEval.}
\label{tab:numpy_eval_score}
\end{table}

\subsection{Case Study on SFT vs. PPO}
\label{appendix:case_study}
\begin{figure*}[ht]
\centering
    \includegraphics[width=1.3\columnwidth]{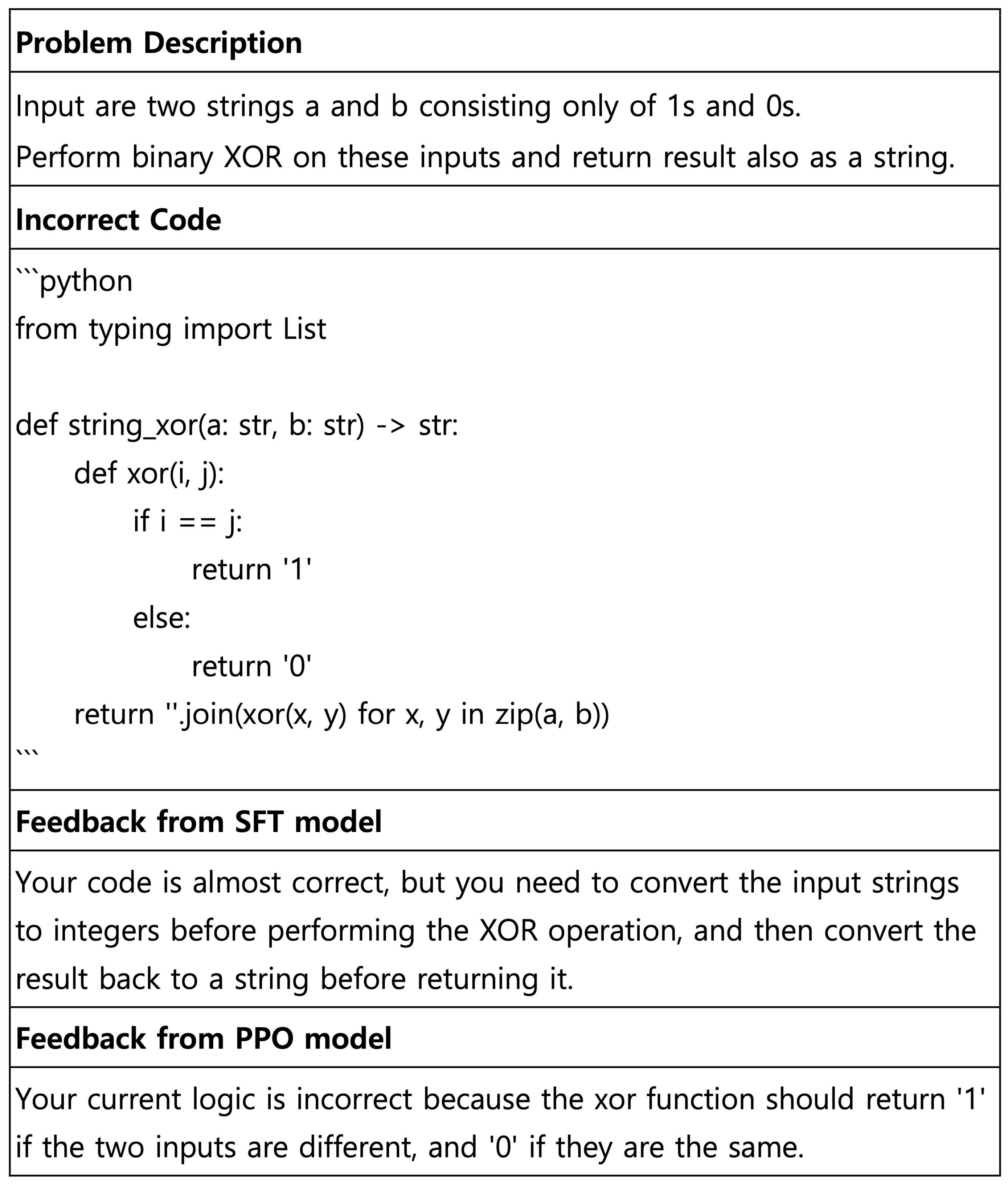}
\caption{Examples of the feedback from SFT and PPO model in \coffeegym.}
\label{fig:case_study_feedback}
\end{figure*}

In Figure~\ref{fig:case_study_feedback}, we present examples of generated feedback.
Although the feedback generated by the SFT model appears plausible, it provides unnecessary feedback which may confuse the editor in feedback-augmented code editing. In contrast, our model (PPO) provides focused and helpful feedback on the incorrect part without unnecessary information. This result aligns with Figure~\ref{fig:error_analysis}, demonstrating that our model generates more accurate and helpful feedback compared to other models.

\section{Prompts for Our Experiments}

\subsection{Correct Feedback Annotation Prompt}\label{appendix:correct_feedback_prompt}
\begin{tcolorbox}[breakable, toprule at break=0pt, bottomrule at break=0pt,colback=white]
\begin{lstlisting}[style=text, columns=fullflexible]
Generate an explanation, analyzation, and plan to generate code prompt for the last task considering the example task instances. Your plan should show enough intermediate reasoning steps towards the answer. Construct the plan as much as you can and describe the logic specifically. When constructing the plan for the code prompt, actively use 'if else statement' to take different reasoning paths based on the condition, 'loop' to efficiently process the repititive instructions, 'dictionary' to keep track of connections between important variables.

[Example 1]
Example task instances:
{example_instances_of_task1}

Output format:
{output_format_of_task1}

Explanation:
{analysis_of_task1}

...

[Example 4]
Example task instances:
{example_instances_of_target_task}

Output format:
{output_format_of_target_task}

Explanation:
\end{lstlisting}
\end{tcolorbox}

\subsection{Wrong Feedback Annotation Prompt}\label{appendix:wrong_feedback_prompt}
\begin{tcolorbox}[breakable, toprule at break=0pt, bottomrule at break=0pt,colback=white]
\begin{lstlisting}[style=text, columns=fullflexible]
Generate feedback that guides the refinement from Code before editing to Code after editing. Assume that the code after editing is 100% correct and your feedback should specifically guide the editing to the code after editing. Please point out only the guidance from the code before editing to the code after editing. Do not provide feedback on the code after editing or any feedback beyond the code after editing.

[Example 1]
Problem Description:
{description}

Code before editing:
{wrong_code}

Code after editing:
{next_wrong_code}

Feedback for Refining the Code:
{feedback}

...

[Example 4]
Problem Description:
{description}

Code before editing:
{wrong_code}

Code after editing:
{next_wrong_code}

Feedback for Refining the Code:
\end{lstlisting}
\end{tcolorbox}

\subsection{Test Case Generation Prompt}\label{appendix:test_case_prompt}
\begin{tcolorbox}[breakable, toprule at break=0pt, bottomrule at break=0pt,colback=white]
\begin{lstlisting}[style=text, columns=fullflexible]
Given the input format and python code, please provide at least 30 challenging test input values to evaluate its functionality.For every start of samples, please attach <start> token to indicate that the input string has started. Also, for every end of samples, please attach <end> token to indicate that the input string has ended.

input format:
{input format}

python code:
{python code}

Sample:
\end{lstlisting}
\end{tcolorbox}

\subsubsection{Code Editor Prompt}\label{appendix:code_editor_prompt}
\begin{tcolorbox}[breakable, toprule at break=0pt, bottomrule at break=0pt,colback=white]
\begin{lstlisting}[style=text, columns=fullflexible]
Provide feedback on the errors in the given code and suggest the correct code to address the described problem.
Description:
{description}
  - output format: {output_format}
  - input format: {input_format}
Incorrect code:
```python
{wrong_code}
```
Feedback:{feedback}

Correct code:
\end{lstlisting}
\end{tcolorbox}

\subsubsection{G-Eval Prompt}
\label{appendix:g_eval}
\begin{tcolorbox}[breakable, toprule at break=0pt, bottomrule at break=0pt,colback=white]
\begin{lstlisting}[style=text, columns=fullflexible]
You will be provided with feedback on the given incorrect code. Classify the accuracy of this feedback using a Likert scale from 1 to 5, where:

1 (Completely incorrect): This feedback has no valid points and is entirely misleading.
2 (Mostly incorrect): This feedback has some valid points but is largely incorrect or misleading.
3 (Neutral or somewhat accurate): This feedback is partially correct but contains significant inaccuracies or omissions.
4 (Mostly correct): This feedback is largely accurate with only minor mistakes or omissions.
5 (Completely correct): This feedback is entirely accurate and provides a correct assessment of the code.
Just generate a score from 1 to 5 based on the accuracy of the feedback.
Description:
{description}
  - output format: {output_format}
  - input format: {input_format}
Incorrect code:
```python
{wrong_code}
```
Feedback:{feedback}

Score:
\end{lstlisting}
\end{tcolorbox}


\end{document}